\documentclass[10pt,twocolumn,letterpaper]{article}

\usepackage{iccv}
\usepackage{makecell}
\usepackage{times}
\usepackage{epsfig}
\usepackage{graphicx}
\usepackage{amsmath}
\usepackage{amssymb}
\usepackage{multirow}
\usepackage{arydshln} 

\usepackage[pagebackref=true,breaklinks=true,letterpaper=true,colorlinks,bookmarks=false]{hyperref}

\iccvfinalcopy 

\def\iccvPaperID{12419} 

\ificcvfinal\fi

\begin{document}

\title{MM-SEAL: A Large-scale Video Dataset of Multi-person Multi-grained Spatio-temporally Action Localization}


\author{Shimin Chen, Wei Li, Chen Chen\\
OPPO Research Institute\\
{\tt\small chenshimin@zju.edu.cn, liwei19@oppo.com}
\and
Jianyang Gu\\
Zhejiang University\\
\and 
Jiaming Chu\\
Beijing University of Posts and Telecommunications\\
\and
Xunqiang Tao, Yandong Guo\\
OPPO Research Institute\\
}

\maketitle
\ificcvfinal\thispagestyle{empty}\fi

\begin{figure*}[t]
\centering
\includegraphics[scale=0.32]{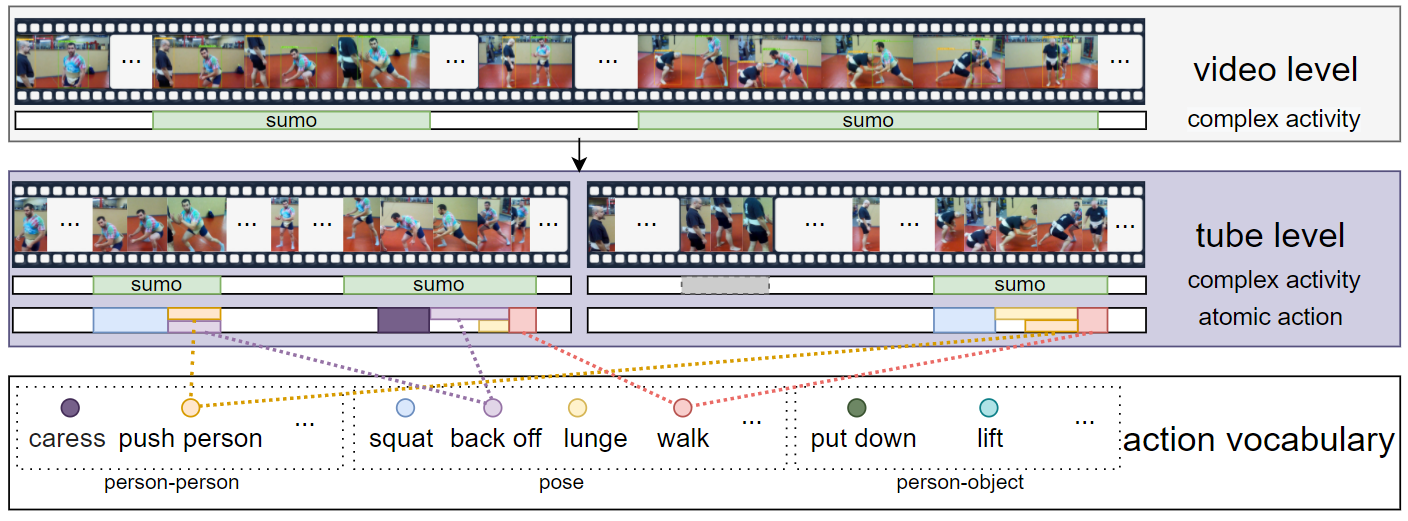}
\caption{\small{The overview of \textit{MM-SEAL} dataset. With the help of multi-object tracking, person re-identification, and annotator correction, we obtain tubelets for each subject who are related to the activity. Then we provide complex activities instances and atomic actions instances of each tubelet. In complex activity level, we annotate instances from 200 classes, such as {``sumo''} marked in green. In atomic action level, we choose atomic actions from the action vocabulary(bottom).}}
\label{fig:overview}
\end{figure*}
\begin{abstract}
   In this paper, we introduce a novel large-scale video dataset dubbed MM-SEAL for multi-person multi-grained spatio-temporal action localization among human daily life. We are the first to propose a new benchmark for multi-person spatio-temporal complex activity localization, where complex semantic and long duration bring new challenges to localization tasks. We observe that limited atomic actions can be combined into many complex activities. MM-SEAL provides both atomic action and complex activity annotations, producing 111.7k atomic actions spanning 172 action categories and 17.7k complex activities spanning 200 activity categories. We explore the relationship between atomic actions and complex activities, finding that atomic action features can improve the complex activity localization performance. Also, we propose a new network which generates temporal proposals and labels simultaneously with adaptively learning the semantic information of proposals, termed Faster-TAD. Finally, our evaluations show that visual features pretrained on MM-SEAL can improve the performance on other action localization benchmarks. We will release the dataset and the project code upon publication of the paper.
\end{abstract}

\section{Introduction}

With the increasing number of videos created every day, video understanding becomes more and more indispensable for the computer vision society. How to locate the spatial positions and temporal action boundaries of multi-person in untrimmed videos is a challenging but meaningful task.

In recent years, action recognition and temporal action detection techniques have gained extraordinary breakthrough due to the advent of many large-scale benchmarks, such as HMDB-51\cite{kuehne2011hmdb}, Kinetics\cite{kay2017kinetics}, ActivityNet\cite{caba2015activitynet} and HACS\cite{zhao2019hacs}. Temporal action detection methods detect actions with boundaries from untrimmed videos, but are unable to spatially detect multiple concurrent human actions. In real-world scenarios, however, it is very common for different people to perform different actions.


Current spatio-temporal localization datasets can be mainly classified into two categories: 1) single-person scene. Datasets such as JHMDB~\cite{jhuang2013towards}, Action Genome\cite{rai2021home} and Homage\cite{rai2021home}. These datasets have promoted impressive progress for spatio-temporal action localization, but fail to deal with multi-person scenes and are limited to the amount of annotations. 2) multi-person scene. AVA~\cite{gu2018ava} involve multi-subject and provide temporal context for labeling the actions in the keyframe of short clips. Their annotations only cover atomic actions with a constant and short temporal range. TITAN\cite{malla2020titan} and MultiSports\cite{li2021multisports} are multi-person datasets and contain fined-grained action categories with dense annotations in both spatial and temporal domains, but they only cover a specific field. Based on the above analysis, there still lacks a large-scale dataset that annotates fine-grained actions in spatial and temporal domains for multi-person scenes in untrimmed daily life videos. On the other hand, the existing spatio-temporal action localization datasets above all aim at actions within 5 seconds, which restricts their application on more complex activities, such as surveillance and healthcare. 

To facilitate the development of video understanding, we introduce a novel large-scale video dataset with multi-person multi-grained spatio-temporal annotations among human daily life, dubbed \textit{MM-SEAL}. We observe that atomic actions can be combined into diverse complex activities. MM-SEAL provides both atomic action and complex activity annotations in tubelet level, producing 111.7k atomic actions spanning 172 action categories and 17.7k complex activities spanning 200 activity categories. As illustrated in Fig.~\ref{fig:overview}, \textit{MM-SEAL Atomic Action} provide fine-grained atomic actions annotations in spatial and temporal dimensions for multi-person in untrimmed daily life videos. Compared with other similar datasets, we not only annotate spatio-temporal action instances, but also provide short-term trajectories for subject person. 
It spurs researches on the relation between the target instance and their contexts like nearby instance of the same person or background scene. \textit{MM-SEAL Complex Activity} are the first to propose a new benchmark for multi-person spatio-temporal complex activity localization, where complex semantic and long duration bring new challenges to spatio-temporal action localization tasks. Compared to datasets (like TITAN, FineGym\cite{shao2020finegym},and MOMA) in the community where both atomic action detection and complex action action detection are proposed, we explore the relationship between atomic actions and complex activities, finding that atomic action features can improve the complex activity localization performance. What's more, we propose a new network for temporal action detection, which generates temporal proposals and action labels simultaneously with adaptively learning the semantic information of proposals, termed Faster-TAD.

Our contributions are summarized as follows: 
\begin{enumerate}
\item We develop a new large-scale benchmark MM-SEAL for multi-person multi-grained spatio-temporal detection in human daily life. 25fps frame-wise annotations for \textit{MM-SEAL Atomic Action}, 4fps frame-wise annotations for \textit{MM-SEAL Complex Activity}.
\item We observe that atomic actions can be combined into diverse complex activities. Thus, we explore the relationship between atomic actions and complex activities, finding that atomic action features can improve the complex activity localization performance. 
\item We propose a baseline for this spatio-temporal localization task. We detect spatial bounding bboxes at the frame-level, and perform temporal action detection. We develop a new network, which generates temporal proposals and action labels simultaneously with adaptively learning the semantic information of proposals, termed Faster-TAD.
\item Our knowledge transfer evaluations show that visual features pretrained on MM-SEAL can improve the performance on other action localization datasets.
\end{enumerate}
\begin{table*}[t]
\centering
\caption{\small{Comparison of statistics between existing action detection datasets and our \textit{MM-SEAL}. (\textit{Keyframe} with action category and spatial localization in keyframe; \textit{Tube} with action category, temporal boundary and spatial localization; \textit{Segment} with action category and temporal boundary; \textit{CA} denotes the \textbf{\textit{C}}omplex \textbf{\textit{A}}ctivity; \textit{AA} denotes the \textbf{\textit{A}}tomic \textbf{\textit{A}}ction; * denotes all action partonomy are count together; Instance means the number of instances)}; \textit{Single/multi} means proving single person or multi person annotations.}
\scalebox{0.89}{
\begin{tabular}{ccccccccc}
\Xhline{2pt}
\multicolumn{1}{c}{Dataset} & \multicolumn{1}{c}{\begin{tabular}[c]{@{}c@{}}Action\\ partonomy\end{tabular}} & \multicolumn{1}{c}{Vid.No} & \multicolumn{1}{c}{Anno type} & \multicolumn{1}{c}{Act.} & \multicolumn{1}{c}{Instance} & \multicolumn{1}{c}{\begin{tabular}[c]{@{}l@{}}avg. act./\\ vid.dur.\end{tabular}} & \multicolumn{1}{c}{Scenes} & \multicolumn{1}{c}{Single/multi} \\ \Xhline{1pt}
 AVA~\cite{gu2018ava} &- & 430  & Key frame   & 80   & 385k     & -                  &human activity & multi \\
AVA-Kinetics~\cite{li2020ava} &- & 239k  & Key frame     & 80   & 624k   & -  & human activity &multi \\  \Xhline{0.3pt} 
ActivityNet-1.3~\cite{caba2015activitynet} &- & 19.99k & Segments     & 200  & 23.1k    & 51.4s/1.9m         & human activity  &-  \\
HACS~\cite{zhao2019hacs}      &-   & 50k    & Segments     & 200  & 140k     & 40.6s/2.6m         & human activity  & -  \\
\hdashline[1pt/5pt]
\multirow{2}{*}{\begin{tabular}[c]{@{}c@{}}FineGym~\cite{shao2020finegym} \end{tabular}}  &action & 303   & Segments     & 10   & 4.9k     & 55s/2h           &Sports &-     \\
&sub-action & 303   & Segments    & 530  & 32.7k    & 1.7s/10m             & Sports  & -\\ \Xhline{0.3pt}
UCF101-24~\cite{singh2017online}  &-  & 15.5k & Tube         & 24   & 4.5k     & 5.1s/6.9s          & human activity &multi\\
JHMDB~\cite{jhuang2013towards} & - & 5.1k &- &21 &0.8k-1.2k & - & human activity & single\\
Multisports~\cite{li2021multisports}  &action & 197.6k & Tube         & 66   & 37.7k    & 1.0s/20.9s         &Sports &multi \\ 
TITAN\cite{malla2020titan} &-&700& Tube & 50 & - & - & egocentric driving&multi\\
MEVA~\cite{corona2021meva} & action & 2.21k & Tube & 37 & 66.2k & -/5m & surveillance & multi\\
\hdashline[1pt/5pt]
\multicolumn{1}{c}{TSU~\cite{dai2022toyota}} & \multicolumn{1}{c}{\begin{tabular}[c]{@{}c@{}}activity \\ atomic action\end{tabular}} & \multicolumn{1}{c}{536*} & \multicolumn{1}{c}{Tube} & \multicolumn{1}{c}{\begin{tabular}[c]{@{}c@{}}5 \\ 46\end{tabular}} & \multicolumn{1}{c}{40.7k*} & 
\multicolumn{1}{c}{-/21m} & smartroom & single\\
\hdashline[1pt/5pt]
\multirow{2}{*}{\begin{tabular}[c]{@{}c@{}}MOMA~\cite{luo2021moma}\end{tabular}}& activity & 2.4k & Segments & 67& -&-& human activity&multi\\ 
& atomic action & 12k & Tube & 52& -&-& human activity&multi\\ \Xhline{0.3pt}
\multirow{2}{*}{\begin{tabular}[c]{@{}c@{}}\textit{MM-SEAL}~\cite{shao2020finegym} 
\end{tabular}}
&activity    & 5.4k  & Tube         & 200  &  17.7k        &  12.04s/1.8m & human activity &multi   \\

&atomic action  & 19.3k     & Tube         & 172  & 111.7k   & 3.62s/9.23s             & human activity &multi   \\ \Xhline{2pt}
\end{tabular}}
\label{fig:table1}
\end{table*}
\section{Related Work}

In recent years, some works have developed action classification \cite{feichtenhofer2019slowfast,christoph2016spatiotemporal,lin2019tsm,qiu2017learning,simonyan2014two,tran2018closer,wang2013action,wang2016temporal} into temporal action localization~\cite{hou2017tube,kalogeiton2017action,singh2017online,wu2020context,yang2019step}, and even spatio-temporal action localization~\cite{ding2018weakly,lea2017temporal,lea2016segmental,lei2018temporal,li2020ms,richard2017weakly,wang2020boundary}. It manifests the trend of video understanding in untrimmed domains.

\subsection{Spatio-temporal Action Localization Datasets.}

A series of datasets, with spatio-temporal annotations, have been introduced in both single-person and multi-person scenarios. For the single-person scenarios, JHMDB~\cite{jhuang2013towards} provide dense spatial localization frame by frame. Action Genome~\cite{ji2020action} decomposes actions into spatio-temporal scene graphs via sparse sampling. Subsequently, HOMAGE~\cite{rai2021home} is proposed, equipped with hierarchical activity and atomic action labels. It also provides multiple viewpoints information and captures object relationships in the scene graph. However, temporal localization in HOMAGE is limited to atomic actions, excluding high-level activities. A recently released dataset, TSU\cite{dai2022toyota} contains dense annotations including elementary, composite activities and activities involving interactions with objects,performed in a spontaneous manner. TSU~\cite{dai2022toyota} is similar to our dataset, but it focuses on smarthome and single-actor scene.

In real-world practical applications, multi-person scenarios are also very common, and a variety of datasets have conducted research in this area. MEVA~\cite{corona2021meva} presents the Multiview Extended Video with spatio-temporal Activities annotations, which aimed at surveillance. The average clip length of MEVA is 5 minutes, but each action is relatively atomic and shorter than MM-SEAL. MOMA~\cite{luo2021moma}, has proposed a redefined action parsing for complex human activity recognition. It organizes action categories with four levels. It is worth noting that neither MultiSports nor MOMA are annotated spatially with complex activities, but only with sub-level actions. 

\subsection{Spatio-temporal action detection method}
Most recent approaches based on datasets UCF101-24 and JHMDB can be divided into two categories: frame-level detectors and clip-level detectors. The frame-level detectors\cite{weinzaepfel2015learning} detect proposals and actions at the frame-level, and then employ the specific linking strategy to generate instance tubes along temporal dimension. However, frame-level detectors fail to fully exploit temporal context for semantic action classification. In contrast, the clip-level detectors model temporal continuity in the videos. Typical researches, ACT\cite{kalogeiton2017action} and MOC-detector\cite{li2020actions}, take a sequence of K frames as input and output K-frame tubelet detection results. The results are linked along temporal dimension into tubes via a common matching strategy. Clip-level detectors can be effectively applied to spatio-temporal localization tasks for atomic actions. However, the number of input frames K, limits the model to capture the features from long-term information. As a result, clip-level detectors can hardly meet the requirements of spatio-temporal localization tasks for activities with complex semantics and long duration. 

\section{The MM-SEAL Dataset}
The purpose of this work is to build a large-scale video dataset for multi-person multi-grained Spatio-tEmporal Action Localization(\textit{MM-SEAL}) among human daily life. In this section, we present the data collection process, the statistics, and the characteristics of \textit{MM-SEAL}.

\subsection{Data Collection \label{Data Collection}}
\subsubsection{Category Selection.}
We follow five principles to generate our atomic action vocabulary. Following the method in AVA\cite{gu2018ava}, generality and exhaustivity are both considered. We collect generic actions in daily life and iterate our action list in several rounds. \textit{MM-SEAL} also follows a principle of fine-grained because some person-object interaction actions have vastly different contexts even within an activity class. For example, there are some activities with intra-class variety such as {``cook’’}. {``cook’’} can be divided into {``wash’’}, {``cut’’}, {``grind’’}, et al. in our dataset. The fourth principle is that our action vocabulary focus on dynamic actions, which means that we only annotate dynamic actions like {``stand up’’} and {``sit down’’}, instead of static pose actions like {``stand’’}, {``sit’’}. We prefer to focus on patterns of motion, which we think is more valuable. For the last principle, the atomic actions should be visible. We end up with 61 pose actions, 15 person-person interaction actions and 96 person-object interaction actions. On the other hand, complex activity annotations adopt the taxonomy of 200 action classes, taken from ActivityNet-1.3\cite{caba2015activitynet}.
\vspace{-0.3cm}
\subsubsection{Data Selection.} 
For the purpose of obtaining atomic actions and complex activities simultaneously, we randomly choose 5,376 videos from ActivityNet-1.3 and HACS. For complex activity detection, we annotate 17,712 complex activity instances spanning 200 categories in 4,224 videos. For atomic action detection, annotators annotate 111,680 atomic actions instances spanning 172 categories within complex activity instances in 5,376 videos. 
\vspace{-0.3cm}
\subsubsection{Subject Selection.} When multiple subjects are present in a video, only main subjects will be annotated in trainset, ignoring background persons. Subjects are selected whose actions are related to complex activities. For efficiency, we select up to three subjects at the same time period. It should be noted that the subject need to be with head shown in at least one-third of frames. Firstly, we get bounding boxes of persons with a detector\cite{tan2020efficientdet}. Top 8 boxes with the highest detection confidence in center areas are selected as subjects. Then, these subjects are feed into MOT algorithm\cite{wojke2017simple}, generating coarse tubelets. We select the top 50 longest tubelets as the candidate subjects. Thirdly, candidate subjects are shown to annotators. Subject selection follows three principles, being in the central area, having high detection confidence, and being related to complex activities. We annotate actions for each person tubelet in raw videos, allowing duration overlap among subject tubelets.

\subsection{Data Annotation}
The action annotations of tubelets are given as a set of bounding boxes in each frame, person ID, start time, end time and action category. In this section, we first introduce the labeling process about tubes. Our annotation team is composed of 11 experienced annotators. The annotators are trained for a week before the formal annotation. To guarantee the quality, each video is labeled by an annotator, reviewed by 2 to 6 annotators. The whole annotation process lasts for 1 year.
\vspace{-0.2cm}
\subsubsection{Spatial Annotation.\label{Spatial Annotation}}
We localize a subject spatially with bounding box and distinguish him or her from other subjects with person ID. In order to effectively obtain accurate spatial annotations, we propose a four-step detection method: 1) Using algorithms to get coarse results; 2) Annotators refine results in short-duration; 3) Utilizing person re-identification technology to merge person IDs in long-duration; 4) Annotators refine results in long-duration. 

Firstly, we adopt a remarkable detector~\cite{tan2020efficientdet} to get bounding boxes of each person. Then, subject selection in frames is proceeded, which is described in Subject Selection Chapter. We generate coarse person IDs using DeepSORT~\cite{wojke2017simple}. In this way, we obtain up to 25 bounding boxes in 1-second of a subject. Secondly, considering that MOT algorithm~\cite{wojke2017simple} leads to ID lost or switch in some cases, annotators are asked to refine it within a short atomic action duration. Notably, we find the detector obtaining accurate positions in most cases, while obvious false alarms are discarded. For efficiency, we refine spatial annotation within short atomic action instance duration. Thirdly, person re-identification technology is utilized to merge tubelets in long-duration. We employ a video re-identification model based on the strong baseline \cite{luo2019strong}. Fourthly, annotators refine results in long-duration. Person IDs that are switched by scene change are not merged.
\vspace{-0.2cm}
\subsubsection{Temporal and Semantic Annotation.}
We provide the start time, end time and semantic labels of action instances. We unify the annotations of repeated actions and actions. For actions that repeat key-moment over a period of time, we define the beginning and ending moment, such as {``cut''}, starting from one second before holding the knife and ending with one second after stopping cutting.
 \begin{figure*}[t]
\centering
\includegraphics[scale=0.42]{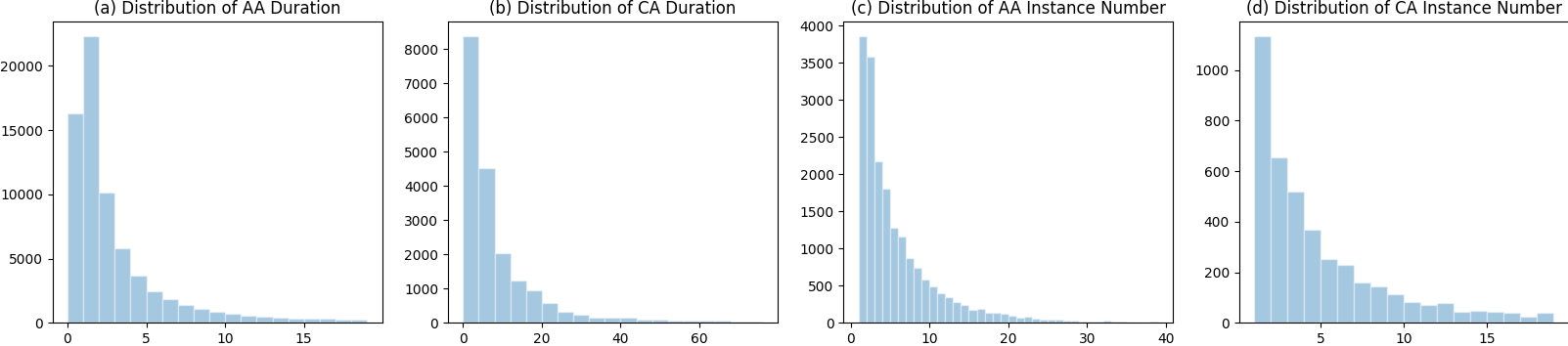}
\caption{\small{The distribution of instance duration(a)(b) and instance numbers in each video(c)(d) for action instances. The abscissa in (a)(b) represents mean the duration of action in seconds, and that in (c)(d) represents the instance numbers in each video. The ordinate represents the instance numbers to the abscissa. \textbf{AA} denotes the \textbf{A}tomic \textbf{A}ction; \textbf{CA} denotes the \textbf{C}omplex \textbf{A}ctivity.}}
\label{fig:hist_segments}
\end{figure*}
\begin{table}[t]
 \caption{\small{Synergy of the actions of different person in the same keyframe in the dataset.}}
   \centering
         \begin{tabular}{l|l|l}
        \Xhline{2pt}
        Person 1 action & Person 2 action & Number  \\ \Xhline{1.2pt}
        shake legs  & beat with hands & 5298 \\ \hline
        wave(object) & bounce & 3825 \\ \hline
        twist waist & dance & 2815 \\ \hline
        lift &  walk & 2277 \\ \hline
        lift & run & 1722 \\ \hline
        step aerobics & rotate & 1587 \\ \hline
        dance & bounce & 1587 \\ \hline
        dance & turn to & 1445 \\ \hline
        rotate & dance & 1441 \\ \hline
        eat & clamp & 879 \\ 
        \Xhline{2pt}
    \end{tabular}
   \label{confuse}
\end{table}

 For complex activity localization, we refer to the boundary annotations in ActivityNet-1.3\cite{caba2015activitynet} and HACS\cite{zhao2019hacs}, and adjust the boundaries of actions for each person. Complex activity semantic annotations adopt the taxonomy of 200 action classes, which are taken from ActivityNet-1.3. For atomic action localization, we annotate atomic action instances within complex activity duration. We propose an action vocabulary for atomic actions, which is described in Data Collection Chapter.

\subsubsection{Quality Assurance.}
This subsection proposes many approaches to assure the quality of \textit{MM-SEAL}. For spatial and temporal annotations, the first priority is to make sure the refined annotation obtained by annotators can be traced in coarse results obtained by the algorithm in forward steps, like step 2 to step 1, step 4 to step 3. Steps are described in {``Spatial Annotation’’} subsection of this Chapter. Then, we propose algorithms to check whether the annotations are out of bounds, and whether the person ID conflicts. Thirdly, in order to check the boundaries for each atomic action instance, we visualize the annotation, which assists annotators to review. Each video will be reviewed by 2 to 6 annotators. 
For semantic annotations, we adopt a cyclic refinement approach including model discrimination, annotator recheck, annotator refinement, ..., model discrimination, and manual recheck.

\subsection{Statistics}
MM-SEAL is composed of 5,376 videos in human daily life with 1/5 for testing and others for training and validation. 

We present a comparison of dataset statistics between \textit{MM-SEAL} and some representative video datasets in Table \ref{fig:table1}. In \textit{MM-SEAL}, there are 17,712 temporal complex activity instances in 4,224 videos and 111,680 atomic action instances in 5,376 videos. Besides, there are 172 atomic action labels defined in this work and 200 complex activity labels which are the same as ActivityNet-1.3. According to statistics, on average, each video contains 4.78 complex action instances, 21.05 atomic action instances and 3.67 atomic action categories. The distribution of instance duration and instance numbers in each video are shown in Fig.~\ref{fig:hist_segments}. The instance duration of atomic actions is significantly less than that of complex activities, yet the instance number of atomic actions is more than that of complex activities. We also present the distribution of atomic action number, shown in Supplementary Materials.


\subsection{Characteristics and Challenges}
There are several characteristics of \textit{MM-SEAL} dataset, which is also challenges on our datast.

Action hierarchy: MM-SEAL contains 172 atomic actions and 200 complex activities. The intra-class variety in dataset with more fine-grained atomic actions, which helps the detection of complex activity.

Long duration: For MM-SEAL AA, We link recurring actions, which results in the prominence of atomic actions with long duration compared with other datasets focused on atomic action(3.62s vs 1.7s on FineGym vs 1.0s on Multiports). For MM-SEAL CA, 1)the average duration of MM-SEAL(12.04s) is 2.36 times that of UCF101-24(5.1s). 2)The mean instance duration of each CA category ranges from 5.4 seconds to 41.0 seconds.
 
Diversity: 1)We annotate three action types: Pose actions, person-person interaction actions and person-object interaction actions. There are 39.75\% of bounding boxes have at least 1 pose action label, 63.63\% of bounding boxes have at least 1 person-object interaction label, which shows that \textit{MM-SEAL} has rich person-object interaction action instances. 2)rich scenarios, we annotate actions in human activity videos rather than a certain scene.

Complex semantics: 1)On average, each video contains 4.78 complex action instances, 21.05 atomic action instances and 3.67 atomic action categories. 2)multi-person. Table \ref{confuse} shows the synergy of the actions of different persons in the same key frame. It demonstrates the diversity of behaviors from different people in the same frame, which manifests that multi-person spatio-temporal action detection is of great significance. 3)Movement. We only annotate dynamic actions. We present a figure in Supplementary Materials to illustrates the distribution of bounding box sizes, showing MM-SEAL contains many boxes with small sizes. Figure illustrates bounding box center offset in one second, showing that 50\% of boxes offset over 50 pixel. Our densely annotation helps to improve the performance of detecting large motion actions.

High temporal variance: 1)The mean instance duration of each atomic action category and complex activity category are shown in Supplementary Materials. The duration of atomic actions ranges from 0.70 seconds to 10.33 seconds, and complex activity ranges from 5.4 seconds to 41.0 seconds. 2)action instances can be long or short(ranges from 0.7s to 233s)

\begin{figure}[t]
\centering
\includegraphics[scale=0.16]{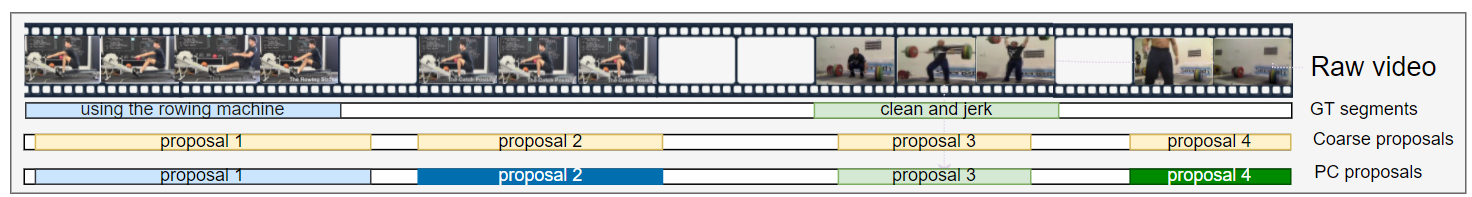}
\caption{\small{Proximity-Category Proposal Block. The first row shows the ground truth segments. The second row is coarse proposals from Proposal Generation Mechanism. The last row shows that the proposal with unsatisfied IoU will be set to a Proximity-Category according to its nearby ground truth segment. For example, proposal 2 has a label of {``using the rowing machine - proximity’’}.}}
\label{proximity-category}
\end{figure}

\begin{figure}
\centering
\includegraphics[scale=0.35]{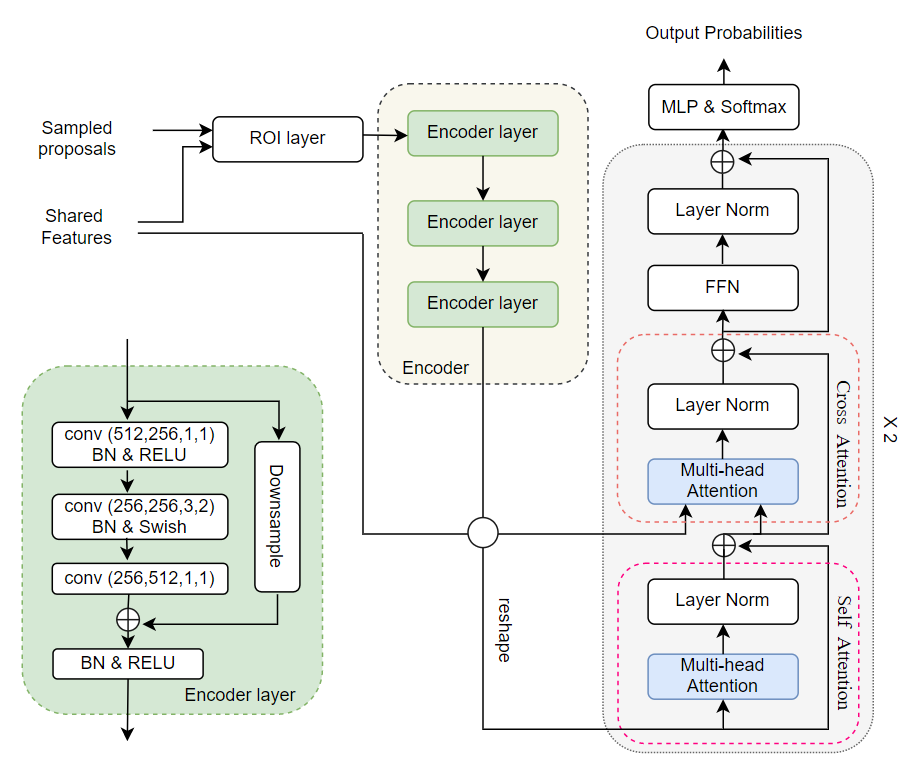}
\caption{\small{Proposal Attention Module. Proposal features are generated from proposal generation outputs and the shared features by a ROI layer. Then, encoder layer is followed to further encode the proposal representation. Finally, Self and Cross Attention block is applied to model the proposal semantic features.}}
\label{fig:classification_head}
\end{figure}

\begin{figure}[t]
\centering
\includegraphics[scale=0.25]{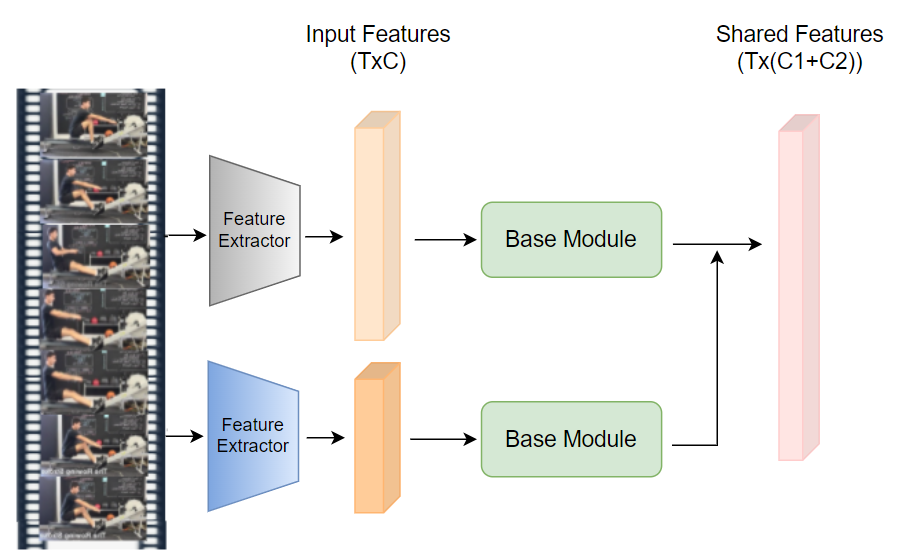}
\caption{\small{Auxiliary-Feature Block. Two streams of features go through base module respectively. Then, they are combined along the temporal dimension. The rest of the network keeps the same.\label{Auxiliary-feature}}}
\vspace{-0.5cm}
\end{figure}

\section{Baseline Approach \label{Baseline Approach}}
\textit{MM-SEAL} develop two benchmarks: atomic spatio-temporal action detection and complex spatio-temporal activity detection. We propose a baseline for this spatio-temporal localization task. We first detect proposals at the frame-level, then track high-scoring proposals throughout the video using a tracking-by-detection approach. The spatio-temporal action localization task is transformed into temporal action localization task. It should be noted that we only perform semantic classification in the temporal action localization stage, effectively using temporal information. Rich labels in our dataset encourage the researchers to consider temporal action proposal and classification in a single framework. 

For the baseline method, we think that the temporal action detection is one of the difficulties. so we focus on this and propose a SOTA network(Faster-TAD). To explore our algorithmic capabilities of temporal action localization, we use gt bboxes as the detected bboxes below. In this way, performance is measured by the average mAP(\%) at different tIoU thresholds(0.5 to 0.95 with 0.05 interval). The metric is employed from \cite{THUMOS14}. Following the standard practice \cite{weinzaepfel2015learning,kalogeiton2017action}, we utilize frame-mAP and video-mAP to evaluate spatio-temporal action detection performance. 

\subsection{Faster-TAD}
Current mainstream approaches ~\cite{lin2018bsn,qing2021temporal,xu2020g} are multi-step solutions which achieve good performance. They include proposal generation, action classification, ensemble results of classifiers and proposal post-processing. However, they fall short in efficiency and flexibility, especially for videos with diverse semantic labels. In recent years, there are also some works focused on single network\cite{lin2021learning,xu2017r}, but they fail to yield comparable results as those of multi-step approaches.

To simplify the pipeline of TAD, we propose a novel single network with remarkable performance, dubbed Faster-TAD. Inspired by Faster-RCNN\cite{ren2015faster}, we jointly learn temporal proposal generation, action classification, and proposal refinement with multi-task loss, sharing information for end-to-end update.

In classification head, we propose a new Context-Adaptive Proposal Module, which consists of Proximity-Category Proposal Block(Fig \ref{proximity-category}), Self-Attention Block, and Cross-Attention Block. Self-Attention Block, and Cross-Attention Block are shown in Fig \ref{fig:classification_head}, which greatly enhance semantic information for proposals. Context-Adaptive Proposal Module is an efficient attention module, which adaptively learn the semantic information through three aspects :a)proposals b)whole video clip feature c)considering context as proximity category proposals.

Many complex human activities have long duration and consist of atomic actions. Action recognition model Swin Transformer\cite{liu2021swin} is adopted to extract features of each clip as input for subsequent localization task. Nevertheless, action recognition model is trained with trimmed short clips. To address this issue, we adopt atomic features as auxiliary features extracted by Slowfast\cite{feichtenhofer2019slowfast} trained on \textit{MM-SEAL Atomic Actions}. We designed a feature aggregation method named \textit{Auxiliary-Features Block} to adapt to the two streams input. As shown in Fig. \ref{Auxiliary-feature}, main and auxiliary features are combined in a simple way after going through two separate base modules.

Extensive experiments demonstrate that Faster-TAD outperforms existing single-network detectors by a large margin on many temporal action detection benchmarks, obtaining state-of-the-art results on ActivityNet-1.3 and SoccerNet-Action Spotting\cite{giancola2018soccernet}. Algorithmic details and experiments are attached in Supplementary Materials.
\vspace{-0.2cm}
\subsection{Atomic spatio-temporal action detection}
We extract features of raw videos utilizing the Slowfast model\cite{feichtenhofer2019slowfast} with $windows=32$, and a $stride=4$. Model is trained on our \textit{MM-SEAL Atomic Action} to get atomic action features. In each window, we use 4fps bounding boxes to extract person features by ROI layer \cite{he2017mask}. The final tubelet feature is termed as Slowfast-T. Faster-TAD is employed to obtain boundaries and semantic labels of action instances for each tubelet. Experimental results are given in Table \ref{table:task23}. We set \textit{MM-SEAL} videos belonging to the training set in ActivityNet-1.3 to the \textit{MM-SEAL} training set. 
\vspace{-0.2cm}
\subsection{Complex spatio-temporal activity detection}
We conduct comparative experiments with two configurations, whose results are shown in Table~\ref{table:task23}. We adopt the released checkpoint of TSP model to extract video features with windows of \textit{size}=16, \textit{stride}=2. In each window, we use 2fps bounding boxes to extract person features by an ROI layer \cite{he2017mask}. The final tubelet feature is termed as TSP-T. This configuration obtains 67.66\% mAP in top120 categories. On the other hand, we adopt Slowfast-T(atomic action feature) as auxiliary features. we concatenate TSP-T and Slowfast-T, obtaining two-stream features as the input for Faster-TAD. Under this configuration, the experiment obtains 68.93\% mAP in top120 categories, bringing a mAP gain of 1.27\%. Experiments demonstrate that learning the features of atomic actions is helpful for complex activity localization task.

\begin{table}[t]
    \centering
    \begin{center}
    \caption{\small{The results of \textit{MM-SEAL Atomic Action} detection(top) and \textit{MM-SEAL Complex Activity} detection(bottom) on validation set, 
   \textit{\textbf{MM-SEAL CA}} denotes the MM-SEAL \textbf{\textit{C}}omplex \textbf{\textit{A}}ctivity. F-mAP means Frame-mAP, V-mAP stands for Video mAP. T-mAP stands for Temporal action detection mAP.}} \label{table:task23}
    \begin{minipage}[t]{1\textwidth}
    \scalebox{0.9}{
    \begin{tabular}{p{2.2cm}|p{1.55cm}|p{1.05cm}|p{1.05cm}|p{1.06cm}}
    \Xhline{2pt}
         Dataset &Features &T-mAP &F-mAP&V-mAP\\ \Xhline{1.5pt}
         MM-SEAL AA &Slowfast-T & 22.82 &20.79&17.55\\ 
        \Xhline{2pt}
    \end{tabular}}
    \end{minipage}
    \\[5pt]
    \begin{minipage}[t]{1\textwidth}
    \scalebox{0.92}{
    
    \begin{tabular}{p{2.3cm}|p{2.7cm}|p{0.6cm}p{0.6cm}p{0.6cm}}
    \Xhline{2pt}
          \multirow{2}{*}{Dataset} & \multirow{2}{*}{Features}  &  \multicolumn{3}{c}{mAP} \\ \cline{3-5} 
         &     & T- & Frame & Video \\ \Xhline{1.2pt}
          MM-SEAL-CA &Slowfast-T&32.09 &44.09& 41.63\\
          MM-SEAL-CA & TSP-T & 67.66 & - & - \\ 
          MM-SEAL-CA & TSP-T+Slowfast-T & 68.93 & - & - \\ 
        \Xhline{2pt}
        
    \end{tabular}}
    \end{minipage}
    \end{center}
\end{table}

\begin{table*}[htbp]
\caption{Comparison of the state-of-the-art methods on MM-SEAL, UCF101-24 and JHMDB.}
 \label{tab:method on CA}
\scalebox{0.92}
\centering
\begin{tabular}{c|c|c|c|c|c|c|c|c|c|c|c|c|c|c}
\Xhline{2pt}
\multicolumn{3}{c|}{\multirow{2}*{Method}} & \multicolumn{3}{c|}{SEAL-CA}& \multicolumn{3}{c|}{SEAL-AA} &\multicolumn{3}{c|}{UCF101-24} & \multicolumn{3}{c}{JHMDB} \\
\cline{4-15}
\multicolumn{3}{c|}{} & $F_{0.5}$ & $V_{0.2}$ & $V_{0.5}$ & $F_{0.5}$ & $V_{0.2}$ & $V_{0.5}$ & $F_{0.5}$ & $V_{0.2}$ & $V_{0.5}$ & $F_{0.5}$ & $V_{0.2}$ & $V_{0.5}$\\
\hline
\multicolumn{3}{c|}{YOWO} & 0.08 & 0.24 & 0.07 & - & - & - & 71.10 & 72.97 & 46.42 & 74.51 & 88.05 & 82.57\\
\multicolumn{3}{c|}{MOC(K=7, top60)} & 0.52 & 1.22 & 0.23 & 13.88 & 3.52 & 0.46 & 78.0 & 82.8 & 53.8 & 70.8 & 77.3 & 77.2\\
\multicolumn{3}{c|}{MOC(K=7, top120)} & 0.22 & 0.93 & 0.51 & 5.78 & 1.47 & 0.19 & - & - & - & - & - & -\\
\multicolumn{3}{c|}{MOC(K=11, top60)} & 1.16 & 2.20 & 0.88 & 15.62 & 3.74 & 0.57 & - & - & - & - & - & - \\
\multicolumn{3}{c|}{MOC(K=11, top120)} & 0.48 & 0.91 & 0.37 & 7.54 & 2.10 & 0.26 & - & - & - & - & - & - \\
\Xhline{1pt}
\end{tabular}
\end{table*}
\vspace{-0.2cm}
\section{Experiments and Analysis}
\subsection{Metrics}
Following the standard practice \cite{weinzaepfel2015learning,kalogeiton2017action}, we utilize frame-mAP and video-mAP to evaluate spatio-temporal action detection performance. IoU at the frame-level is adopted to evaluate frame-mAP, for which the threshold is 0.5. Similarly, video-mAP is calculated by IoU between two tubes. The threshold is 0.2 and 0.5. 
\subsection{Results and Analysis}
 We evaluate several typical action detection frameworks on MM-SEAL, and compare their performance on UCF101-24 and JHMDB.

As shown in Table\ref{tab:method on CA}, MOC-detector and YOWO achieve excellent results in Both UCF101-24 and JHMDB, while getting a poor performance on our MM-SEAL CA. YOWO and MOC-Detector predict bounding boxes and action probabilities directly from video clips. However, they perform poorly when detecting an activity with a long duration which need a large receptive field temporally. On the other hand, compared with other datasets, MM-SEAL contain more complex semantics and more precise temporal annotations. K is defined as number of frames fed to model. We set K as 7 and 11 separately, and find a improvement in performance as K increased.

Unlike UCF101-24, MM-SEAL provide 4fs annotations because of the long duration of complex activity. Each frame is fed into the MOC detector to learn moving point trajectories by estimating movement at adjacent frames on UCF101-24. For complex activity, we feed the model 4fs video frames, which we guess greatly affects the results. Comparing the above method with our baseline, we observe that how to connect trajectories is very important in spatio-temporal action detection tasks with long instance duration.

For our atomic actions, we observe that the frame-MAP of top60(13.88\% k=7) is much larger than the frame-mAP of top120(5.78\% k=7). The long tail effect of the number of instances between categories has a great influence on the results. UCF101-24 and JHMDB have only the same label of activity for each video, which provides enough characteristic backgroud cues for detectors. Meanwhile, MM-SEAL provide multiple label of actions within a video. Furthermore, MM-SEAL involves concurrent actions for one person and different people, which bring many challenges for this task.

\section{Knowledge Transfer}
To demonstrate the effectiveness of \textit{MM-SEAL} dataset, we have done several spatio-temporal localization knowledge transfer experiments.

\subsection{Atomic Spatio-temporal Action Localization}
The detailed atomic action annotations possess \textit{MM-SEAL} with strong generalization capability over other action localization tasks. For example, recent works usually adopt the backbone model pre-trained on Kinetics-700~\cite{carreira2017quo} for downstream tasks fine-tuning. Kinetics-700 is a large-scale dataset which focuses on action recognition. We hope that by sharing the same target on the atomic action localization task, \textit{MM-SEAL} can play a more active role in improving the model performance on AVA. 

We validate the generalization capability of our proposed \textit{MM-SEAL} dataset to AVA in Table~\ref{tab:trans_ava}. By simply conducting pretraining, \textit{MM-SEAL} can provide a much better initialization for AVA fine-tuning. To better utilize the consistent target of atomic action localization, we propose to conduct a semi-supervised adaptation process to help the model pre-trained on \textit{MM-SEAL Atomic Action} adapt to AVA dataset. Algorithmic details are attached in Supplementary Materials.
The results manifest that assigning proper pseudo labels for AVA allows the model to build better priors for the subsequent fine-tuning. 

\subsection{Temporal Action Localization}
In recent years, researchers have found that the representational capability of input features is very important for the localization task. For example, TSP \cite{alwassel_2021_tsp} is an approach focusing on temporally-sensitive pre-training of video encoders. It is observed that atomic actions can be combined into diverse complex activities. We explore the relationship between atomic actions and complex activities by applying atomic action features extrated from raw video to complex activity detection tasks.

\begin{table}[t]
    \centering
    \begin{center}
    \caption{\small{Performance comparison of initialization methods on AVA fine-tuning.}}
    \label{tab:trans_ava}
    \scalebox{0.92}{
    \begin{tabular}{p{2.3cm}|p{4cm}|p{1.3cm}}
    \Xhline{2pt}
        Method & Pre-train Dataset & $T_{50}$ mAP \\ \Xhline{1.2pt}
         Pre-train & Kinetics-700 & 29.3 \\
        Pre-train & \textit{MM-SEAL Atomic Action} & 30.4 \\ \Xhline{1.2pt}
        Semi-Transfer & \textit{MM-SEAL Atomic Action} & 30.8 \\ \Xhline{2pt}
    \end{tabular}}
    \end{center}
\end{table}

\begin{table}[t]\small
\centering
\caption{\small{Action detection results on validation set of ActivityNet-1.3 and HACS, measured by mAP(\%) at different tIoU thresholds and the average mAP(\%) at different tIoU thresholds(0.5 to 0.95 with 0.05 interval). \textit{SF-A} means feature extracted by Slowfast \cite{feichtenhofer2019slowfast} trained on our \textit{MM-SEAL Atomic Actions}.}}
\scalebox{0.94}{
\begin{tabular}{p{1.1cm}|p{1.6cm}|p{1.2cm}|p{1.4cm}|p{1.3cm}} 
\Xhline{2pt}
Datasets & Feature & 0.5@mAP & 0.95@mAP & Avg mAP \\ 
\Xhline{1.2pt}
\multirow{4}{*}{ANet-1.3} & TSP & 51.29 & 10.22 & 35.32 \\ 
                        & TSP+SF-A& 52.20 & 10.10& \textbf{35.98} \\ 
                            & Swin                     & 57.39                    & 10.48                     & 39.09                     \\ 
                            & Swin+SF-A            & 58.30                     & 11.28                     & \textbf{40.01}                     \\ 
\hline
\multirow{2}{*}{HACS}       & Swin                     & 54.13               & 12.02                     & 36.92                     \\ 
                            & Swin+SF-A            & 55.63                    & 12.90                     & \textbf{38.39}                     \\ 
\Xhline{2pt}
\end{tabular}}
\label{The result of ActivityNet-1.3 and Hacs instance.}
\end{table}

Feature extracted by Slowfast \cite{feichtenhofer2019slowfast} trained on our \textit{MM-SEAL Atomic Actions} is named as Slowfast-A Feature. Swin Feature indicates features extracted by the Swin-Transformer~\cite{liu2021swin} trained on HACS Clips and TSP Feature is trained on ActivityNet-1.3. We adopt Faster-TAD in this task. Slowfast-A Feature is employed as auxiliary features and assembled with TSP Feature or Swin Feature, generating two-stream inputs for localization task. 

ActivityNet-1.3 and HACS are commonly adopted to evaluate the capabilities of algorithms on temporally localizing activities in untrimmed video sequences. Extensive experiments are employed on these benchmark, and results are shown in Table \ref{The result of ActivityNet-1.3 and Hacs instance.}. We can see that visual features trained on MM-SEAL can improve the performance of other temporal action localization task. What's more, atomic action features can improve the complex activity localization performance.
\vspace{-0.2cm}
\section{Conclusions}
We develop a new large-scale benchmark MM-SEAL for multi-person multi-grained spatio-temporal detection among human daily life. We are the first to propose a new benchmark for multi-person spatio-temporal complex activity localization, where complex semantic and long duration bring new challenges to video understanding. We observe that atomic actions can be combined into diverse complex activities, and prove the great effect of the atomic features on complex activity localization task. Also, we propose a baseline method equipped with novel network Faster-TAD and hope our \textit{MM-SEAL} will spur researches on spatio-temporal action localization tasks. Finally, Our evaluations show that visual features pretrained on MM-SEAL can improve the performance on other action localization benchmarks.

{\small
\bibliographystyle{ieee_fullname}
\bibliography{egbib}

\begin{thebibliography}{10}\itemsep=-1pt

\bibitem{alwassel_2021_tsp}
Humam Alwassel, Silvio Giancola, and Bernard Ghanem.
\newblock Tsp: Temporally-sensitive pretraining of video encoders for
  localization tasks.
\newblock In {\em Proceedings of the IEEE/CVF International Conference on
  Computer Vision (ICCV) Workshops}, 2021.

\bibitem{caba2015activitynet}
Fabian Caba~Heilbron, Victor Escorcia, Bernard Ghanem, and Juan Carlos~Niebles.
\newblock Activitynet: A large-scale video benchmark for human activity
  understanding.
\newblock In {\em Proceedings of the ieee conference on computer vision and
  pattern recognition}, pages 961--970, 2015.

\bibitem{carreira2017quo}
Joao Carreira and Andrew Zisserman.
\newblock Quo vadis, action recognition? a new model and the kinetics dataset.
\newblock In {\em proceedings of the IEEE Conference on Computer Vision and
  Pattern Recognition}, pages 6299--6308, 2017.

\bibitem{christoph2016spatiotemporal}
R Christoph and Feichtenhofer~Axel Pinz.
\newblock Spatiotemporal residual networks for video action recognition.
\newblock {\em Advances in neural information processing systems}, pages
  3468--3476, 2016.

\bibitem{corona2021meva}
Kellie Corona, Katie Osterdahl, Roderic Collins, and Anthony Hoogs.
\newblock Meva: A large-scale multiview, multimodal video dataset for activity
  detection.
\newblock In {\em Proceedings of the IEEE/CVF Winter Conference on Applications
  of Computer Vision}, pages 1060--1068, 2021.

\bibitem{dai2022toyota}
Rui Dai, Srijan Das, Saurav Sharma, Luca Minciullo, Lorenzo Garattoni, Francois
  Bremond, and Gianpiero Francesca.
\newblock Toyota smarthome untrimmed: Real-world untrimmed videos for activity
  detection.
\newblock {\em IEEE Transactions on Pattern Analysis and Machine Intelligence},
  45(2):2533--2550, 2022.

\bibitem{ding2018weakly}
Li Ding and Chenliang Xu.
\newblock Weakly-supervised action segmentation with iterative soft boundary
  assignment.
\newblock In {\em Proceedings of the IEEE Conference on Computer Vision and
  Pattern Recognition}, pages 6508--6516, 2018.

\bibitem{feichtenhofer2019slowfast}
Christoph Feichtenhofer, Haoqi Fan, Jitendra Malik, and Kaiming He.
\newblock Slowfast networks for video recognition.
\newblock In {\em Proceedings of the IEEE/CVF international conference on
  computer vision}, pages 6202--6211, 2019.

\bibitem{giancola2018soccernet}
Silvio Giancola, Mohieddine Amine, Tarek Dghaily, and Bernard Ghanem.
\newblock Soccernet: A scalable dataset for action spotting in soccer videos.
\newblock In {\em Proceedings of the IEEE conference on computer vision and
  pattern recognition workshops}, pages 1711--1721, 2018.

\bibitem{gu2018ava}
Chunhui Gu, Chen Sun, David~A Ross, Carl Vondrick, Caroline Pantofaru, Yeqing
  Li, Sudheendra Vijayanarasimhan, George Toderici, Susanna Ricco, Rahul
  Sukthankar, et~al.
\newblock Ava: A video dataset of spatio-temporally localized atomic visual
  actions.
\newblock In {\em Proceedings of the IEEE Conference on Computer Vision and
  Pattern Recognition}, pages 6047--6056, 2018.

\bibitem{he2017mask}
Kaiming He, Georgia Gkioxari, Piotr Doll{\'a}r, and Ross Girshick.
\newblock Mask r-cnn.
\newblock In {\em Proceedings of the IEEE international conference on computer
  vision}, pages 2961--2969, 2017.

\bibitem{hou2017tube}
Rui Hou, Chen Chen, and Mubarak Shah.
\newblock Tube convolutional neural network (t-cnn) for action detection in
  videos.
\newblock In {\em Proceedings of the IEEE international conference on computer
  vision}, pages 5822--5831, 2017.

\bibitem{jhuang2013towards}
Hueihan Jhuang, Juergen Gall, Silvia Zuffi, Cordelia Schmid, and Michael~J
  Black.
\newblock Towards understanding action recognition.
\newblock In {\em Proceedings of the IEEE international conference on computer
  vision}, pages 3192--3199, 2013.

\bibitem{ji2020action}
Jingwei Ji, Ranjay Krishna, Li Fei-Fei, and Juan~Carlos Niebles.
\newblock Action genome: Actions as compositions of spatio-temporal scene
  graphs.
\newblock In {\em Proceedings of the IEEE/CVF Conference on Computer Vision and
  Pattern Recognition}, pages 10236--10247, 2020.

\bibitem{THUMOS14}
Y.-G. Jiang, J. Liu, A. Roshan~Zamir, G. Toderici, I. Laptev, M. Shah, and R.
  Sukthankar.
\newblock {THUMOS} challenge: Action recognition with a large number of
  classes.
\newblock \url{http://crcv.ucf.edu/THUMOS14/}, 2014.

\bibitem{kalogeiton2017action}
Vicky Kalogeiton, Philippe Weinzaepfel, Vittorio Ferrari, and Cordelia Schmid.
\newblock Action tubelet detector for spatio-temporal action localization.
\newblock In {\em Proceedings of the IEEE International Conference on Computer
  Vision}, pages 4405--4413, 2017.

\bibitem{kay2017kinetics}
Will Kay, Joao Carreira, Karen Simonyan, Brian Zhang, Chloe Hillier, Sudheendra
  Vijayanarasimhan, Fabio Viola, Tim Green, Trevor Back, Paul Natsev, et~al.
\newblock The kinetics human action video dataset.
\newblock {\em arXiv preprint arXiv:1705.06950}, 2017.

\bibitem{kuehne2011hmdb}
Hildegard Kuehne, Hueihan Jhuang, Est{\'\i}baliz Garrote, Tomaso Poggio, and
  Thomas Serre.
\newblock Hmdb: a large video database for human motion recognition.
\newblock In {\em 2011 International conference on computer vision}, pages
  2556--2563. IEEE, 2011.

\bibitem{lea2017temporal}
Colin Lea, Michael~D Flynn, Rene Vidal, Austin Reiter, and Gregory~D Hager.
\newblock Temporal convolutional networks for action segmentation and
  detection.
\newblock In {\em proceedings of the IEEE Conference on Computer Vision and
  Pattern Recognition}, pages 156--165, 2017.

\bibitem{lea2016segmental}
Colin Lea, Austin Reiter, Ren{\'e} Vidal, and Gregory~D Hager.
\newblock Segmental spatiotemporal cnns for fine-grained action segmentation.
\newblock In {\em European Conference on Computer Vision}, pages 36--52.
  Springer, 2016.

\bibitem{lei2018temporal}
Peng Lei and Sinisa Todorovic.
\newblock Temporal deformable residual networks for action segmentation in
  videos.
\newblock In {\em Proceedings of the IEEE conference on computer vision and
  pattern recognition}, pages 6742--6751, 2018.

\bibitem{li2020ava}
Ang Li, Meghana Thotakuri, David~A Ross, Jo{\~a}o Carreira, Alexander
  Vostrikov, and Andrew Zisserman.
\newblock The ava-kinetics localized human actions video dataset.
\newblock {\em arXiv preprint arXiv:2005.00214}, 2020.

\bibitem{li2020ms}
Shi-Jie Li, Yazan AbuFarha, Yun Liu, Ming-Ming Cheng, and Juergen Gall.
\newblock Ms-tcn++: Multi-stage temporal convolutional network for action
  segmentation.
\newblock {\em IEEE transactions on pattern analysis and machine intelligence},
  2020.

\bibitem{li2021multisports}
Yixuan Li, Lei Chen, Runyu He, Zhenzhi Wang, Gangshan Wu, and Limin Wang.
\newblock Multisports: A multi-person video dataset of spatio-temporally
  localized sports actions.
\newblock In {\em Proceedings of the IEEE/CVF International Conference on
  Computer Vision}, pages 13536--13545, 2021.

\bibitem{li2020actions}
Yixuan Li, Zixu Wang, Limin Wang, and Gangshan Wu.
\newblock Actions as moving points.
\newblock In {\em European Conference on Computer Vision}, pages 68--84.
  Springer, 2020.

\bibitem{lin2021learning}
Chuming Lin, Chengming Xu, Donghao Luo, Yabiao Wang, Ying Tai, Chengjie Wang,
  Jilin Li, Feiyue Huang, and Yanwei Fu.
\newblock Learning salient boundary feature for anchor-free temporal action
  localization.
\newblock In {\em Proceedings of the IEEE/CVF Conference on Computer Vision and
  Pattern Recognition}, pages 3320--3329, 2021.

\bibitem{lin2019tsm}
Ji Lin, Chuang Gan, and Song Han.
\newblock Tsm: Temporal shift module for efficient video understanding.
\newblock In {\em Proceedings of the IEEE/CVF International Conference on
  Computer Vision}, pages 7083--7093, 2019.

\bibitem{lin2018bsn}
Tianwei Lin, Xu Zhao, Haisheng Su, Chongjing Wang, and Ming Yang.
\newblock Bsn: Boundary sensitive network for temporal action proposal
  generation.
\newblock In {\em Proceedings of the European conference on computer vision
  (ECCV)}, pages 3--19, 2018.

\bibitem{liu2021swin}
Ze Liu, Yutong Lin, Yue Cao, Han Hu, Yixuan Wei, Zheng Zhang, Stephen Lin, and
  Baining Guo.
\newblock Swin transformer: Hierarchical vision transformer using shifted
  windows.
\newblock In {\em Proceedings of the IEEE/CVF International Conference on
  Computer Vision}, pages 10012--10022, 2021.

\bibitem{luo2019strong}
Hao Luo, Wei Jiang, Youzhi Gu, Fuxu Liu, Xingyu Liao, Shenqi Lai, and Jianyang
  Gu.
\newblock A strong baseline and batch normalization neck for deep person
  re-identification.
\newblock {\em IEEE Transactions on Multimedia}, 22(10):2597--2609, 2019.

\bibitem{luo2021moma}
Zelun Luo, Wanze Xie, Siddharth Kapoor, Yiyun Liang, Michael Cooper,
  Juan~Carlos Niebles, Ehsan Adeli, and Fei-Fei Li.
\newblock Moma: Multi-object multi-actor activity parsing.
\newblock {\em Advances in Neural Information Processing Systems},
  34:17939--17955, 2021.

\bibitem{malla2020titan}
Srikanth Malla, Behzad Dariush, and Chiho Choi.
\newblock Titan: Future forecast using action priors.
\newblock In {\em Proceedings of the IEEE/CVF Conference on Computer Vision and
  Pattern Recognition}, pages 11186--11196, 2020.

\bibitem{qing2021temporal}
Zhiwu Qing, Haisheng Su, Weihao Gan, Dongliang Wang, Wei Wu, Xiang Wang, Yu
  Qiao, Junjie Yan, Changxin Gao, and Nong Sang.
\newblock Temporal context aggregation network for temporal action proposal
  refinement.
\newblock In {\em Proceedings of the IEEE/CVF Conference on Computer Vision and
  Pattern Recognition}, pages 485--494, 2021.

\bibitem{qiu2017learning}
Zhaofan Qiu, Ting Yao, and Tao Mei.
\newblock Learning spatio-temporal representation with pseudo-3d residual
  networks.
\newblock In {\em proceedings of the IEEE International Conference on Computer
  Vision}, pages 5533--5541, 2017.

\bibitem{rai2021home}
Nishant Rai, Haofeng Chen, Jingwei Ji, Rishi Desai, Kazuki Kozuka, Shun
  Ishizaka, Ehsan Adeli, and Juan~Carlos Niebles.
\newblock Home action genome: Cooperative compositional action understanding.
\newblock In {\em Proceedings of the IEEE/CVF Conference on Computer Vision and
  Pattern Recognition}, pages 11184--11193, 2021.

\bibitem{ren2015faster}
Shaoqing Ren, Kaiming He, Ross Girshick, and Jian Sun.
\newblock Faster r-cnn: Towards real-time object detection with region proposal
  networks.
\newblock {\em Advances in neural information processing systems}, 28, 2015.

\bibitem{richard2017weakly}
Alexander Richard, Hilde Kuehne, and Juergen Gall.
\newblock Weakly supervised action learning with rnn based fine-to-coarse
  modeling.
\newblock In {\em Proceedings of the IEEE conference on Computer Vision and
  Pattern Recognition}, pages 754--763, 2017.

\bibitem{shao2020finegym}
Dian Shao, Yue Zhao, Bo Dai, and Dahua Lin.
\newblock Finegym: A hierarchical video dataset for fine-grained action
  understanding.
\newblock In {\em Proceedings of the IEEE/CVF conference on computer vision and
  pattern recognition}, pages 2616--2625, 2020.

\bibitem{simonyan2014two}
Karen Simonyan and Andrew Zisserman.
\newblock Two-stream convolutional networks for action recognition in videos.
\newblock {\em Advances in neural information processing systems}, 27, 2014.

\bibitem{singh2017online}
Gurkirt Singh, Suman Saha, Michael Sapienza, Philip~HS Torr, and Fabio
  Cuzzolin.
\newblock Online real-time multiple spatiotemporal action localisation and
  prediction.
\newblock In {\em Proceedings of the IEEE International Conference on Computer
  Vision}, pages 3637--3646, 2017.

\bibitem{tan2020efficientdet}
Mingxing Tan, Ruoming Pang, and Quoc~V Le.
\newblock Efficientdet: Scalable and efficient object detection.
\newblock In {\em Proceedings of the IEEE/CVF conference on computer vision and
  pattern recognition}, pages 10781--10790, 2020.

\bibitem{tran2018closer}
Du Tran, Heng Wang, Lorenzo Torresani, Jamie Ray, Yann LeCun, and Manohar
  Paluri.
\newblock A closer look at spatiotemporal convolutions for action recognition.
\newblock In {\em Proceedings of the IEEE conference on Computer Vision and
  Pattern Recognition}, pages 6450--6459, 2018.

\bibitem{wang2013action}
Heng Wang and Cordelia Schmid.
\newblock Action recognition with improved trajectories.
\newblock In {\em Proceedings of the IEEE international conference on computer
  vision}, pages 3551--3558, 2013.

\bibitem{wang2016temporal}
Limin Wang, Yuanjun Xiong, Zhe Wang, Yu Qiao, Dahua Lin, Xiaoou Tang, and
  Luc~Van Gool.
\newblock Temporal segment networks: Towards good practices for deep action
  recognition.
\newblock In {\em European conference on computer vision}, pages 20--36.
  Springer, 2016.

\bibitem{wang2020boundary}
Zhenzhi Wang, Ziteng Gao, Limin Wang, Zhifeng Li, and Gangshan Wu.
\newblock Boundary-aware cascade networks for temporal action segmentation.
\newblock In {\em European Conference on Computer Vision}, pages 34--51.
  Springer, 2020.

\bibitem{weinzaepfel2015learning}
Philippe Weinzaepfel, Zaid Harchaoui, and Cordelia Schmid.
\newblock Learning to track for spatio-temporal action localization.
\newblock In {\em Proceedings of the IEEE international conference on computer
  vision}, pages 3164--3172, 2015.

\bibitem{wojke2017simple}
Nicolai Wojke, Alex Bewley, and Dietrich Paulus.
\newblock Simple online and realtime tracking with a deep association metric.
\newblock In {\em 2017 IEEE international conference on image processing
  (ICIP)}, pages 3645--3649. IEEE, 2017.

\bibitem{wu2020context}
Jianchao Wu, Zhanghui Kuang, Limin Wang, Wayne Zhang, and Gangshan Wu.
\newblock Context-aware rcnn: A baseline for action detection in videos.
\newblock In {\em European Conference on Computer Vision}, pages 440--456.
  Springer, 2020.

\bibitem{xu2017r}
Huijuan Xu, Abir Das, and Kate Saenko.
\newblock R-c3d: Region convolutional 3d network for temporal activity
  detection.
\newblock In {\em Proceedings of the IEEE international conference on computer
  vision}, pages 5783--5792, 2017.

\bibitem{xu2020g}
Mengmeng Xu, Chen Zhao, David~S Rojas, Ali Thabet, and Bernard Ghanem.
\newblock G-tad: Sub-graph localization for temporal action detection.
\newblock In {\em Proceedings of the IEEE/CVF Conference on Computer Vision and
  Pattern Recognition}, pages 10156--10165, 2020.

\bibitem{yang2019step}
Xitong Yang, Xiaodong Yang, Ming-Yu Liu, Fanyi Xiao, Larry~S Davis, and Jan
  Kautz.
\newblock Step: Spatio-temporal progressive learning for video action
  detection.
\newblock In {\em Proceedings of the IEEE/CVF Conference on Computer Vision and
  Pattern Recognition}, pages 264--272, 2019.

\bibitem{zhao2019hacs}
Hang Zhao, Antonio Torralba, Lorenzo Torresani, and Zhicheng Yan.
\newblock Hacs: Human action clips and segments dataset for recognition and
  temporal localization.
\newblock In {\em Proceedings of the IEEE/CVF International Conference on
  Computer Vision}, pages 8668--8678, 2019.

\end{thebibliography}
}

\end{document}


\title{Supplementary Material for MM-SEAL: A Large-scale Video Dataset of Multi-person Multi-grained Spatio-temporally Action Localization}

\author{First Author\\
Institution1\\
Institution1 address\\
{\tt\small firstauthor@i1.org}
\and
Second Author\\
Institution2\\
First line of institution2 address\\
{\tt\small secondauthor@i2.org}
}

\maketitle
\ificcvfinal\thispagestyle{empty}\fi

\begin{abstract}
   In this supplementary material, we present fully detailed information on 1) MM-SEAL Statistics; 2) MM-SEAL Characteristics; 3) Quality Assurance for semantic annotations; 4) Faster-TAD; 5) Our semi-supervised adaptation process; 6) Data format; 7) Atomic action vocabulary.
\end{abstract}

\section{MM-SEAL Statistis}
This appendix provides figures to present statistis of MM-SEAL. Fig\ref{fig:action_number} presents the distribution of total seconds of each atomic action class annotated. Fig\ref{fig:figure2} shows the mean instance duration of each atomic action category and complex activity category. We can see that the duration of atomic actions ranges from 0.70 seconds to 10.33 seconds, which is shorter than that of complex activity ranging from 5.4 seconds to 41.0 seconds.

\section{MM-SEAL Characteristics}

This appendix explores more characteristics of \textit{MM-SEAL} dataset. We adopt the \textit{SEAL-Atomic Action} to make sufficient statistics on the synergy of the actions and the characteristic of boxes in the dataset. Fig.~\ref{fig:hist_clips_0301}(a) illustrates the distribution of bounding box sizes. There are still many boxes with small sizes. Fig.~\ref{fig:hist_clips_0301}(b) illustrates bounding box center offset in one second. Dense annotation helps to improve the detection performance of actions with movement. Table~\ref{confuse} shows the synergy of the actions of the same person in the same key frame. The person with lifting action are always with others walking.


 \begin{table}
   \centering
         \begin{tabular}{l|l|l} \Xhline{2pt}
        Action 1 & Action 2 & Number  \\ \Xhline{1.2pt}
        lift & walk & 8391 \\ \hline
        wave(object) & bounce & 2735 \\ \hline
        lift & run & 1881 \\ \hline
        push(object) & walk & 1855 \\ \hline
        lift & stand up & 1339 \\  \hline
        stand up & flat press & 1148 \\  \hline
        shake legs & beat with object & 1134 \\  \hline
        lift & turn to & 1011 \\  \hline
        put down & lift & 830 \\  \hline
        Play skateboarding & bounce & 786 \\  \hline
        turn to & walk & 776 \\  \hline
        wave ones' hand & twist ones' waist & 749\\\Xhline{2pt}
    \end{tabular}
    \caption{\small{Synergy of the actions of different person in the same keyframe in the dataset.}}
   \label{confuse}
\end{table}

\section{Quality Assurance for semantic annotations}
This appendix describes in detail our quality assurance for semantic annotations. We adopt a cyclic refinement approach including model discrimination, annotator re-check, annotator refinement, ..., model discrimination, and manual recheck. It is notable that model discrimination is only used for quality assurance, not for annotating. Firstly, when the annotation amount reaches half of the total data, we use a baseline approach \cite{tang2020asynchronous} to classify actions, and check the accuracy of action category annotations. Top 5 mis-classified actions are picked out to be analyzed individually. Then, we judge the action confusion according to the similarity of action patterns. The action categories with identical semantics are merged. Secondly, annotators correct the results according to the refined categories. Thirdly, we use new annotations to train models. These steps are looped for multiple times.
\begin{figure*}[t]
\centering
\includegraphics[scale=0.28]{iccv2023AuthorKit/action_number_20230303_hacs_add_old.png}
\caption{Total seconds of each atomic action class annotated in the \textit{MM-SEAL Atomic Action} sorted by descending order, with colors indicating action types.}
\label{fig:action_number}
\end{figure*}

\begin{figure}
\centering
\includegraphics[scale=0.4]{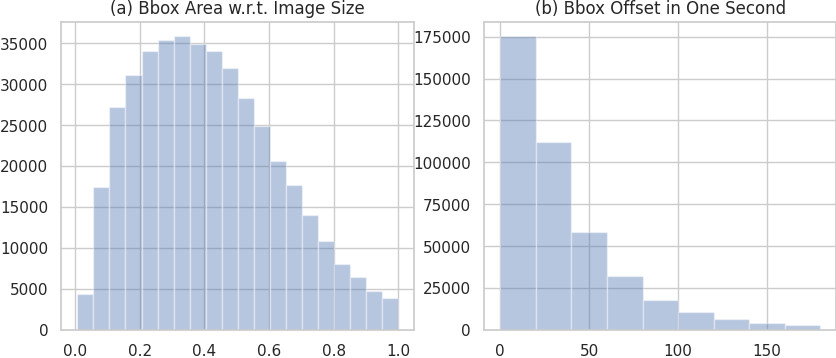}
\caption{The distribution of bounding box size and bounding box offset.}
\label{fig:hist_clips_0301}
\vspace{-0.5cm}
\end{figure}
\section{Faster-TAD}
To simplify the pipeline of TAD, we propose a novel single network with remarkable performance, dubbed Faster-TAD. Inspired by Faster-RCNN\cite{ren2015faster}, we jointly learn temporal proposal generation, action classification, and proposal refinement with multi-task loss, sharing information for end-to-end update.

We observe many challenges in temporal action detection. Firstly, to tackle the extreme duration variation of action instances, ranging from second to minute, boundary-based mechanism\cite{lin2019bmn} is adopted to generate proposals instead of the traditional anchor-based method. Secondly, proposal context is helpful for the recognition of the proposal label. To enhance semantic information for action instances, we bring an innovative Context-Adaptive Proposal Module for classification, in which we propose a new Proximity-Category Proposal Block to obtain context, a Self-Attention Block to construct relationships among proposals, and a Cross-Attention Block to learn relevant context in raw videos for proposals. Thirdly, we propose a new Fake-
Proposal Block to make refinement module to be trained with various offsets relative to ground truth boundary. Last but not least, many complex human activities have long duration and consist of atomic actions, so action recognition models such as CSN\cite{tran2019video} are often adopted to extract the features of video clips as input for subsequent localization task. Nevertheless, action recognition model trained with limited input frames and complex human activity labels lacks atomic action information which may improve the recognition of boundaries and classes. To address this issue, we utilize atomic features trained on \textit{MM-SEAL Atomic Action} as auxiliary features.

\begin{figure*}
\centering
\includegraphics[height=5cm,width=12cm]{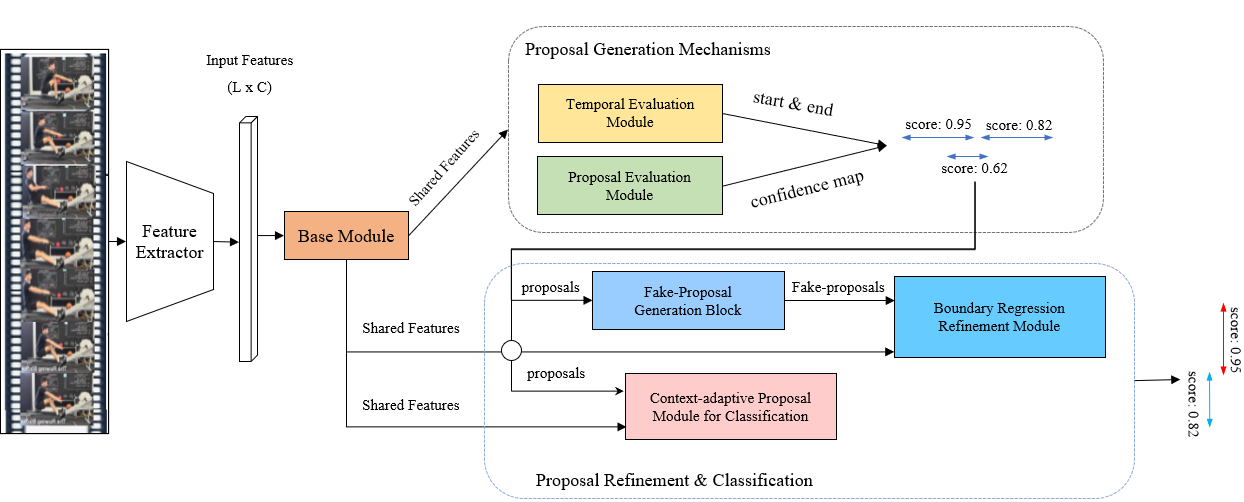}
\caption{Overview of our method. Given an untrimmed video, Faster-TAD can generate proposals and simultaneously (1) refine the boundary and (2) classify the proposal in a context-adaptive way. We construct our Faster-TAD with feature sequences extracted from raw video as inputs.}
\label{figure:faster-tad framework}
\end{figure*}

\subsection{Overview of Framework}
As shown in Fig.~\ref{figure:faster-tad framework}, we propose a Faster-RCNN like network in temporal action detection, Faster-TAD. By jointing temporal proposal generation and action classification with multi-task loss and shared features, Faster-TAD simplifies the pipeline of TAD.

The input to our pipeline is a video sequence $X=\{x_t\}_{t=1}^T$, which can be represented by $X\in\mathbb{R}^{C\times L}$, where $C$ is the feature dimension of each clip, and $L$ is the number of clips. 

We first process the feature sequences with a base module to extract shared features, which consists of a CNN Layer, a Relu Layer, and a GCNeXt\cite{xu2020g} Block. We then exert a Proposal Generation Mechanism to obtain most credible $K$ coarse proposals, where $K$ is 120. Proposals and shared features are further utilized to get more accurate boundaries by Boundary Regression Refinement Module\cite{qing2021temporal}. At the same time, shared features and proposals are employed to get the semantic labels of action instances with Context-Adaptive Proposal Module. We make some improvements to tackle the challenges in temporal action detection. 

\subsection{Context-Adaptive Proposal Module}

In object detection task, there is limited relation between objects in different spaces. However, in temporal action detection task, action instance is closely related to other actions in the same video. For example, if a semantic label of proposal is high jumping, the preceding action is likely to be running. The context greatly helps to classify the semantic label of proposals. We construct this module with Proximity-Category Proposal Block, Self-Attention Block, and Cross-Attention Block, learning useful context adaptively for each proposal.

Our Context-adaptive Proposal Module is illustrated in Fig.4 of main body of our paper. Proposal features are extracted from shared features by the ROI\cite{he2017mask} layer with sampled proposals. Compared with shared features ($X\in\mathbb{R}^{C\times L}$), Proposal features can be represented by $F_p\in\mathbb{R}^{N\times C\times T}$, where $N$ is the number of coarse sampled proposals, $C$ is the feature dimension of each clip, and $T$ is temporal resolution processed by ROI layer. The method to sample proposals is described in {``Proximity-Category Proposal Block’’} subsection. Encoder consists of three encoder layers to obtain deeper semantic features for each proposal. We employ Residual Block\cite{he2016deep} as the encoder layer. After three encoder layers, the temporal dimension of proposal features turns into $1/8 \times T$. Proposal features are afterwards flattened along the last two dimensions to a feature sequence ($P\in\mathbb{R}^{N\times \frac{T}8C})$, where $T$ is set to 16.

\subsection{Proximity-Category Proposals Block}
Previous classification methods generally regard the proposal with IoU larger than a threshold as positive proposals, and other proposals as negative proposals. IoU is the matching score between the proposal and ground truth(GT).

In general, there are two methods to classify proposals\cite{liu2016ssd,redmon2016you,ren2015faster}. The first is to only reduce classification loss for positive proposals, the other is to add a negative category and assign all negative proposals to the negative category. However, negative proposals also contain many semantic information, and it isn't a sensible approach to classify all negative proposals into one category. As illustrated in Fig.3 of main body of paper, we propose a new proposal selection approach, called Proximity-Category Proposals Block. We define proposals with proximity category as Proximity-Category Proposals(PC Proposals). For example, in Fig.3 of main body of our paper, proposal 2 is a PC Proposal of {``using the rowing machine - proximity’’}, and proposal 4 is a PC proposal of {``clean and jerk - proximity’’}. In this way, we expand the origin $M$ categories into $2M$ categories. As shown in algorithm \ref{alg:algorithm}, we propose a high-low threshold method to set semantic labels for each proposal, according to the maximum IoU value between the proposal and ground true segments.

\begin{algorithm}[tb]
\caption{A algorithm to set semantic labels for each proposal}
\label{alg:algorithm}
\textbf{Input}: i, which is a index range from 0 to 2M-1\\
\textbf{Parameter}: ${\tau}_1$, ${\tau}_2$, $\alpha$\\
\textbf{Output}: The semantic label for each proposal, termed as L which is a numpy array of with 1 rows and 2M columns.
\begin{algorithmic}[1] 
\STATE Let $idx=G(argmax(IoUs))$
\STATE Let $idy=G(argmin(Dists))$
\STATE Let each element of L is set to zero.
\IF {$IoU_{best}\geq{\tau}_1$}
\STATE $L(i=idx)=1.0$.
\ELSIF {${\tau}_1>IoU_{best}\geq{{\tau}_2}$}
\STATE $L(i=idx)=\alpha IoU_{best}$
\STATE $L(i=(idx+M))=1-\alpha IoU_{best}$
\ELSE
\STATE $L(i=(idy+M))=1.0$
\ENDIF
\STATE \textbf{return} L
\end{algorithmic}
\end{algorithm}

As illustrated in algorithm \ref{alg:algorithm}, ${\tau}_1$ is the high threshold of $IoU$, ${\tau}_2$ is the low threshold. $M$ is the number of original categories, $i$ is ranging frame $0$ to $2M-1$ in Eq \ref{alg:algorithm}. ${IoUs}$ are the IoU values between proposal and the ground truth segments. ${IoU_{best}}$ is the max value of ${IoUs}$.  $idx$ and $idy$ are the index of label for proposal, $(idx+M)$ and $(idy+M)$ are index of Proximity-Category label for proposal. $\alpha$ is a hyperparameter. As illustrated in algorithm \ref{alg:algorithm}, $G$ is a function that maps ground truth position to ground truth label index. ${Dists}$ stand for the center point distance values between proposal and the ground truth segments.

In order to make the ratio of positive proposals and PC Proposals close to 1:1, we select positive proposals first, and then select PC Proposals with the highest scores until $N$ coarse proposals are sampled.

 \begin{table*}[t]
    \centering
    \scalebox{0.92}{
    \begin{tabular}{c|c|c|c|c|c|c|c}
    \hline
        Method & Feature & S-N & Ensemble of classifiers & 0.5 & 0.75 & 0.95 & Avg  \\\hline
        \multicolumn{8}{l}{Self-contained methods}\\\hline
        R-C3D\cite{xu2017r} & C3D & $\checkmark$ & $\times$ & 26.8 & - & - & - \\
        TAL-Net\cite{chao2018rethinking} & I3D & $\checkmark$ & $\times$ & 38.23 & 18.30 & 1.30 & 20.22  \\
        P-GCN\cite{zeng2019graph} & I3D & $\times$ & $\times$ & 42.90 & 28.14 & 2.47 & 26.99  \\
        TadTR\cite{liu2021end} & I3D & $\checkmark$ & $\times$ & 40.85 & 28.44 & 7.84 & 27.75 \\
        ContextLoc\cite{zhu2021enriching} & I3D &  $\checkmark$ & $\times$ & 51.24 & 31.40 & 2.83 & 30.59\\
        Lin et al.\cite{lin2021learning} & I3D & \checkmark & $\times$ & 52.4 & 35.3 & 6.5 & \textbf{34.4} \\
        \hline 
        \multicolumn{8}{l}{Ensemble Action-labels}\\\hline 
        P-GCN\cite{zeng2019graph} & I3D & $\times$ & UntrimmedNet & 48.26 & 33.16 & 3.27 & 31.11  \\ 
        BMN\cite{lin2019bmn} & TS & $\times$ & CUHK & 50.07 & 34.78 & 8.29 & 33.85  \\ 
        G-TAD\cite{xu2020g} & TS & $\times$ & CUHK & 50.36 & 34.60 & 9.02 & 34.09  \\
        TSP\cite{alwassel2021tsp} & TSP & $\times$ & CUHK & 51.26 & 37.12 & 9.29 & 35.81 \\
        BMN-CSA\cite{sridhar2021class} & TSP & $\times$ & CUHK & 52.64 & 37.75 & 7.94 & 36.25 \\
        TCANet[BMN]\cite{qing2021temporal} & Slowfast &$\times$ & CUHK & 54.33 & 39.13 &8.41 & 37.56\\ 
        PRN\cite{wang2021proposal} & CSN & $\times$ & PRN & 57.9 & - & - & \textbf{39.4} \\\hline
        \multicolumn{8}{l}{Our method}\\\hline
        Faster-TAD & TSP & \checkmark & $\times$ & 51.29 & 36.19 & 10.22 & 35.32\\
        Faster-TAD & TSP+Slowfast-A & \checkmark & $\times$ & 52.20 & 36.97 & 10.10 & 35.98 \\
        Faster-TAD & Swin & \checkmark & $\times$ & 57.39 & 39.97 & 10.48 & \textbf{39.09} \\
        Faster-TAD & Swin+Slowfast-A & \checkmark & $\times$ & 58.30 & 40.77&11.28 & \textbf{40.01} \\\hline
    \end{tabular}}
    \caption{Action detection results on validation set of ActivityNet-1.3, measured by mAP(\%) at different tIoU thresholds and the average mAP(\%). $S-N$ stands for Single-Network. $Ensemble \ of \ classifiers$ stands for ensemble video level classification results. $CUHK$ is from\cite{xiong2016cuhk}, $UntrimmedNet$ is from \cite{wang2017untrimmednets}. $Slowfast-A$ is extracted by Slowfast model trained on \textit{MM-SEAL Atomic Actions}. {``-’’} indicates unknown results.}
    \label{tab:tab1}
\end{table*}

\subsection{Self-Attention Block.}
We utilize Self-Attention\cite{vaswani2017attention} Block to capture relationships between proposals. Our decoder consists of 2 decoder blocks. At each block $l$, Query/Key/Value sequences are computed for proposal features from the representation ${P}^l$ encoder by the preceding block:
\begin{gather}
    {Q}^l={K}^l={V}^l={P}^l 
\end{gather}
Where $l$ is the block index. The multi-head variant of the self-attention computation is popularly used because of jointly attention to information from different representation sub-spaces. Multi-Head Attention\cite{vaswani2017attention} uses scaled dot-product attention, which is defined as $MultiHead-Attn(Q^l,K^l,V^l)$.

We set the LayerNorm\cite{ba2016layer} after the Multi-Head Attention layer, followed with residual connection. We call this approach Middle-LN Transformer Layer, which is defined as:
\begin{equation}
    {P_{out}}^l=LN(MultiHead-Attn({Q}^l,{K}^l,{V}^l))+{Q}^l
\end{equation}

\subsection{Cross-Attention Block.}
We permute the shared features at the dimension of 0 and 1, which is encoded from raw video features by base module, and set the permuted features as key sequences and value sequences. It is represented by Eq (\ref{eq:cross-attn}). That is to say, Cross-Attention Block attends to learn the relationship between proposals and every clips in raw video. This block greatly increases the semantic information of proposals, bringing a large performance improvement.
\begin{equation}
    {K_{cross}}^l={V_{cross}}^l=permute(X),
    {Q_{cross}}^l={P_{out}}^l \label{eq:cross-attn}
\end{equation}
where $X\in\mathbb{R}^{C\times L}$ is shared features, $K_{cross}\in\mathbb{R}^{L\times C}$ is key sequences, $V_{cross}\in\mathbb{R}^{L\times C}$ is value sequences, $C$ is the feature dimension of each clip, and $L$ is the number of clips.

Like Self-Attention Block, we utilize Middle-LN Transformer Layer to learn relationships, which is defined as ${P_{cross}}^l$. ${P_{cross}}^l$ is further employed to get semantic features of proposals by a two-layer Feed-Forward Network(FFN)\cite{vaswani2017attention}, which is defined as:
\begin{equation}
    P^{l+1}=FFN({P_{cross}}^l)
\end{equation}

\begin{figure}[t]
\centering
\includegraphics[scale=0.5]{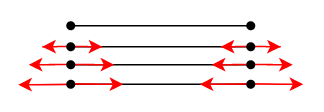}
\caption{Fake-Proposal Block. The red arrows indicate the different offsets based on the origin boundary of an ground truth segment. This block generates $[0,\pm\frac{1}{8},\pm\frac{1}{6},\pm\frac{1}{4}]$ seven modes to offset for each boundary of GT.}
\label{Fake-Proposal}
\end{figure}

\subsection{Fake-Proposal Block}
 We propose a gt-based Fake-Proposal Block in Boundary Regression Refinement Module. We select the $\mu$ proposals with the highest confidence from $K$ coarse proposals as part of the proposals that will be refined. Then, we generate the rest $K-\mu$ fake proposals based on gt. As illustrated in Fig. \ref{Fake-Proposal}, we have $[0,\pm\frac{1}{8},\pm\frac{1}{6},\pm\frac{1}{4}]$ seven modes to change boundaries of GT. That is to say, each GT can generate 49 fake proposals. For each GT, we choose extension ways in order to generate fake proposals until $K-\mu$ fake proposals are generated, where $K$ and $\mu$ are set to 120 and 90.
 
The Fake-Proposal Block ensures that the refinement module can be trained with various boundary offsets relative to GT, which is beneficial for regression. On the other 

\begin{table*}
    \centering
    \scalebox{0.95}{
    \begin{tabular}{c|c|c|c|c|c|c|c}
    \hline
        Method & Feature & S-N & Ensemble of classifiers & 0.5 & 0.75 & 0.95 & Avg  \\\hline
        TadTR\cite{liu2021end} & I3D & $\checkmark$ & $\times$ & 45.16 & 30.70 & 11.78 & 30.83 \\
        G-TAD\cite{xu2020g}& I3D & $\times$ & - & 41.08 & 27.59 & 8.34 & 27.48  \\
        BMN\cite{lin2019bmn} & Slowfast & $\times$ & - & 52.49 & 36.38 & 10.37 & 35.76  \\ 
        TCANet[BMN]\cite{qing2021temporal} & Slowfast &$\times$ & - & 56.74 & 41.14 &12.15 & \textbf{39.77}\\ 
        \hline
        Faster-TAD(ours) & Swin & \checkmark & $\times$ & 54.13 & 37.10 & 12.02 & 36.92 \\
        Faster-TAD(ours) & Swin+Slowfast-A & \checkmark & $\times$ & 55.63 & 38.72 & 12.90 & \textbf{38.39} \\\hline
    \end{tabular}}
    \caption{Action detection results on validation set of HACS-Segments, measured by mAP(\%) at different tIoU thresholds and the average mAP(\%). $S-N$ stands for Single-Network. $Ensemble \ of \ classifiers$ stands for ensemble video level classification results. $Slowfast-A$ is extracted by Slowfast model trained on \textit{MM-SEAL Atomic Actions}. {``-’’} indicates unknown results. Results of BMN are from \cite{qing2021temporal}.}
   \label{tab:tab2}
\end{table*}

\begin{table*}
    \centering
    \scalebox{1.0}{
    \begin{tabular}{c|c|c|c|c}
    \hline
        Method & Feature & Shown & Unshown & All \\\hline
        Zhao et al.\cite{zhou2021feature}(2021 top1) & BaiDu & 52.33&25.63&47.05 \\
        Faster-TAD(ours) & Swin(3s) & 56.91 & 24.38 & 50.34\\
        Faster-TAD(ours) & Swin(3s+6s) & 61.10 & 25.50 & 54.09 \\\hline
    \end{tabular}}
    \caption{Action detection results on test set of SoccerNet-Action Spotting, measured by tight average-mAP introduced by \cite{giancola2018soccernet}. The test set is divided to shown part and unshown part according to the action visibility. We report the performance on Shown, Unshown and all test set. Validation set is defined as test, and test set is called challenge in SoccerNet.}\label{tab:tab3}
\end{table*}

\begin{table}
    \centering
    \begin{center}
    \begin{tabular}{p{2.5cm}|p{2cm}|p{2cm}}
    \Xhline{2pt}
        Method & $\lambda_c$ & mAP(top50) \\ \Xhline{1.2pt}
        Pre-train & - & 30.4 \\
        Semi-Transfer & 0.25 & 30.5 \\
        Semi-Transfer & 0.30 & 30.8 \\
        Semi-Transfer & 0.35 & 30.8 \\ \Xhline{2pt}
    \end{tabular}
    \caption{ Performance comparison of different $\lambda_c$ on AVA fine-tuning. }
    \label{tab:trans_ava}
    \end{center}
\end{table}

\noindent hand, this block feeds more valid proposals to the refinement module.

\subsection{Auxiliary-Features Block}
 This block is described in detail in the paper, and will not be repeated here.

\subsection{Experiment}
We train our model in a single network, with batch size of 64 on 8 gpus. The learning rate is $6\times{10}^{-4}$ for the first 3 epochs, and is reduced by 10 in epoch 3 and 7. We train the model with total 10 epochs. In inference, we apply Soft-NMS\cite{bodla2017soft} for post-processing, and select the top-M prediction for final evaluation. M is 120 for both ActivityNet-1.3 and Hacs Segments. We do not adjust hyper-parameters for HACS Segments, using the same hyper-parameters as those on ActivityNet-1.3.

\textbf{ActivityNet-1.3\cite{caba2015activitynet}}
Table \ref{tab:tab1} demonstrates the temporal action detection performance comparison on validation set of ActivityNet-1.3. Faster-TAD reports the highest average mAP results on this large-scale dataset. Our approach outperforms existing single-network detector by a large margin of 5.61\% mAP. Our single network outperforms these multi-step method by 0.61\% mAP, even multi-step detector using ensemble results of classifiers. 

\textbf{HACS Segments\cite{zhao2019hacs}}
Table \ref{tab:tab2} compares Faster-TAD with state-of-the-art detectors on HACS Segments. Our approach obtains remarkable performance of 38.39\% average mAP, and outperforms existing single-network detector by a large maigin of 7.56\% mAP.

\textbf{SoccerNet-Action Spotting\cite{giancola2018soccernet}}
\cite{zhou2021feature} is the winner of SoccerNet-Action Spotting 2021. As shown in Table \ref{tab:tab3}, with Faster-TAD network, we reached an mAP of 54.09\% in tight test set, bringing a gain of 8.77\% mAP in shown set.

\section{Our semi-supervised adaptation process}
This appendix describes detailedly our semi-supervised adaptation process, which helps the model pre-trained on \textit{SEAL-Atomic Action} adapt to AVA dataset. Different from some previous semi-supervised methods~\cite{ding2021kfc,rashid2020action,singh2021semi}, there are significant differences on the annotation scales between \textit{SEAL-Atomic Action} and AVA. Hence, we employ model pre-trained on \textit{SEAL-Atomic Action} to predict pseudo labels on AVA, and design a sample selection mechanism to filter credible samples. The selection mechanism is summarized as the following two rules. Firstly, the prediction result of the corresponding sample must be discriminate enough, which requires the maximum classification score is larger than a credible threshold $\lambda_c$. Secondly, the prediction result should be robust, for which we apply two different random transforms to the data. The sample is selected only if the predicted class for both inputs are consistent. In addition, to dynamically adjust the label precision and training sample scales, we alternatively conduct the pseudo label predicting and training process.

We study the influence of different sample selection threshold $\lambda_c$ on the semi-supervised transfer in Tab.~\ref{tab:trans_ava}. A proper threshold ensures the credibility of predicted pseudo labels and the number of training samples simultaneously. When the threshold becomes over-low, the pseudo labels gradually include noises, which reduce the effect of the transfer process. But generally the transfer results are robust to fluctuations of $\lambda_c$, and are higher than naive pre-training. 

\begin{listing}[tb]%
\caption{Atomic Spatio-temporal Action Annotations}%
\label{lst:atomic annotations}%
\begin{lstlisting}[language=Haskell]
Tublets with pkl format:
{Video_name:{
    TubletID: tube_boxes, #25fps
    ...
    }
}
Action instances with json format:
{TubeletID:{
    "annotations":[
        {
        "label": atomic-action,
        "segment":[t_start,t_end], 
        },
        {
        ...
        }
      ],
    "durations": duration,
    "fps":fps,
    "url":url,
    "available_areas":[ 
        t1_start,t1_end] 
    }
    ...
}
\end{lstlisting}
\end{listing}

\section{Data format}
We describe the data format of our two tasks in this section, consisting atomic spatio-temporal action annotations, and complex spatio-temporal activity annotations.

\subsection{Atomic Spatio-temporal Action detection.} 
We provide two kinds of annotations: Video instance annotations and Tubelet instance annotations.

1)Video instance annotations

We provide gt annotations, whose format is the same with UCF101-24~\cite{singh2017online} and JHMDB~\cite{jhuang2013towards}. Our gt annotations contain 5 items as follows:
\begin{itemize}
\item[\bullet] labels (list): List of the 172 labels.
\item[\bullet]gttubes (dict): Dictionary that contains the ground truth tubes for each video. A gttube is dictionary that associates with each index of label and a list of tubes. A tube is a numpy array with nframes rows and 5 columns, and each col is in format like [frame index, x1, y1, x2, y2].
\item[\bullet]nframes (dict): Dictionary that contains the number of frames for each video.
\item[\bullet]train\_videos (list): A list with $nsplits=1$ elements, each one containing the list of training videos.
\item[\bullet]test\_videos (list): A list with $nsplits=1$ elements, each one containing the list of testing videos.
\end{itemize}

2)Tubelet instance annotations

We provide tublets with pkl format, which is a dictionary that associates from each video name, a dict of tubes. Tube\_boxes is a numpy array with tube-length rows and 5 columns, every row is [frame number,x1,y1,x2,y2]. Also, we provide action instances with json format, which is a dictionary that associate from each TubeletID. The annotation structure is shown in listing \ref{lst:atomic annotations}.

\subsection{Complex Spatio-temporal Activity Detetcion.} 
We provide two kinds of annotations: Video instance annotations and Tubelet instance annotations.

1)Video instance annotations

We provide gt annotations, whose format is similar to UCF101-24~\cite{singh2017online} and JHMDB~\cite{jhuang2013towards}. Our gt annotations contain 5 items as follows:
\begin{itemize}
\item[\bullet] labels (list): List of the 200 labels.
\item[\bullet]gttubes (dict): Dictionary that contains the ground truth tubes for each video. A gttube is dictionary that associates with each index of label and a list of tubes. A tube is a numpy array with nframes rows and 5 columns, and each col is in format like [frame index, x1, y1, x2, y2]. We provide 4 frame index for each second of video.
\item[\bullet]nframes (dict): Dictionary that contains the number of frames for each video.
\item[\bullet]train\_videos (list): A list with $nsplits=1$ elements, each one containing the list of training videos.
\item[\bullet]test\_videos (list): A list with $nsplits=1$ elements, each one containing the list of testing videos.
\end{itemize}

2)Tubelet instance annotations
We provide 4fps tublets with pkl format, which is a dictionary that associates from each video name, a dict of tubes.The annotation structure is similar to listing \ref{lst:atomic annotations}.
\begin{table}
\centering
\begin{tabular}{p{2cm}p{2cm}p{2.5cm}}
 \Xhline{2pt}
\multicolumn{3}{c}{Person-Person Interaction}           \\  \Xhline{1.2pt}
pull(sb.) & lift(sb.) & talk to(sb.)   \\
push         & give to      & hit           \\
wrestle      & take from    & clash         \\
shake hands  & kick         & arm wrestling \\
hug          & hold down    & block(sb.) \\  \Xhline{2pt}
\end{tabular}
\caption{15 Person-Person actions.}\label{table2}
\end{table}

\begin{table}
\centering
\scalebox{0.92}{
\begin{tabular}{p{2.0cm}p{2.7cm}p{2.8cm}}
\Xhline{2pt}
\multicolumn{3}{c}{Pose}                                 \\ \Xhline{1.2pt}
walk         & climb              & sharpen              \\
back off     & applause           & shake legs           \\
run          & step over          & somersault           \\
bounce       & look back          & guessing game        \\
sit down     & point to           & mark time            \\
kneel down   & wave ones' hand    & twist ones' waist    \\
crouch       & high knees         & punch                \\
stand up     & wipe               & kick                 \\
bend over    & rub hands together & walk on stilts       \\
handstand    & wash ones' hands   & dive                 \\
lie down     & stretch out        & bar muscle up        \\
get up       & shake ones' head   & shake ones' hip      \\
turn to      & pose               & jump off             \\
dance        & pat                & arched back          \\
swim         & blow               & pull                 \\
fall         & tickle             & step aerobics        \\
lunge        & braided hairstyles & leg swing            \\
sit-ups      & wash ones' face    & Thomas Full Spin     \\
flat press   & skate              & raise leg straightly \\
rotate       & jumping jack       & do the splits        \\
bar dip      &                    &                      \\ \Xhline{2pt}
\end{tabular}}
\caption{61 Pose actions.}\label{table3}
\end{table}

\begin{table}[t]
\centering
\scalebox{0.92}{
\begin{tabular}{p{3.9cm}p{1.8cm}p{1.8cm}}
 \Xhline{2pt}
\multicolumn{3}{c}{Person-Object Interaction}                                                         \\ \Xhline{1.2pt}
 swing    & shoot                  & splash                              \\
ride a horse               & weld         & stick-up                            \\
   paddle & shoulder               & put down                            \\
beat sth with object     & ride a bike   & pull                            \\
 shoot a basket          & shave                  & throw                               \\
  knit                    & pluck                  & unpack                              \\
descend the steps             & saute               &pole vault                   \\
drive                      & wave                   & press                               \\
trim vegetables                  & wash sth               &surf                     \\
   glide                  & hang                   & support                             \\
beat sth with hands            & brush sth              & bend                                \\
go up the steps              & pour                   & smoke                     \\
 trample                       & dribble                & wear                                \\
sharpen                    & grind                  & rollover                            \\
use rowing machine         & write                  & fiddle                              \\
shoot an arrow             & clamp                  & move                                \\
walk on narrow objects   & pull out & dip                                 \\
carve              & push                   & sprinkle                            \\
eat                        & smear                  & tear                                \\
Play skateboarding         & dig                    & extruding                           \\
climb on horizontal hanger    & smooth           & knead          \\
walk the dog               & twining                & undress                             \\
play wind instruments                    & ironing                & pounce                              \\
twirl baton                & scoop                  & caress                              \\
play plucked instrument                      & lift                   & roll up                             \\
blow                       & burn                   & uncover                             \\
sweep                      & cut                    & flatten                             \\
pack stuffing to food     & twist(object)          & hack \\
drink                      & shovel                 & snatch                              \\
spray                      & peel                   & poke                                \\
 break off                      & wipe(object)           &  chop                          \\
untie                      & fill in                & stick in                            \\ \Xhline{2pt}
\end{tabular}}
\caption{96 Person-Object actions.}\label{table1}
\end{table}

\section{Atomic action vocabulary}
In this section, we will present all of our atomic action categories, which are divided into ``pose actions’’, ``person-person actions’’ and ``person-object actions’’ according to the interaction object. Table \ref{table3} shows 61 pose actions, Table \ref{table2} shows 15 person-person interaction actions, and Table \ref{table1} shows 96 person-object interaction actions.

\begin{figure*}[ht]
 \centering
  \subfigure[Duration of each atomic action]
  {\includegraphics[scale=0.2]{iccv2023AuthorKit/seg_atomic_20230303_hacs_add_old.png}}\\ 
 \centering
 \subfigure[Duration of each complex activity]
 {\includegraphics[width=17cm]{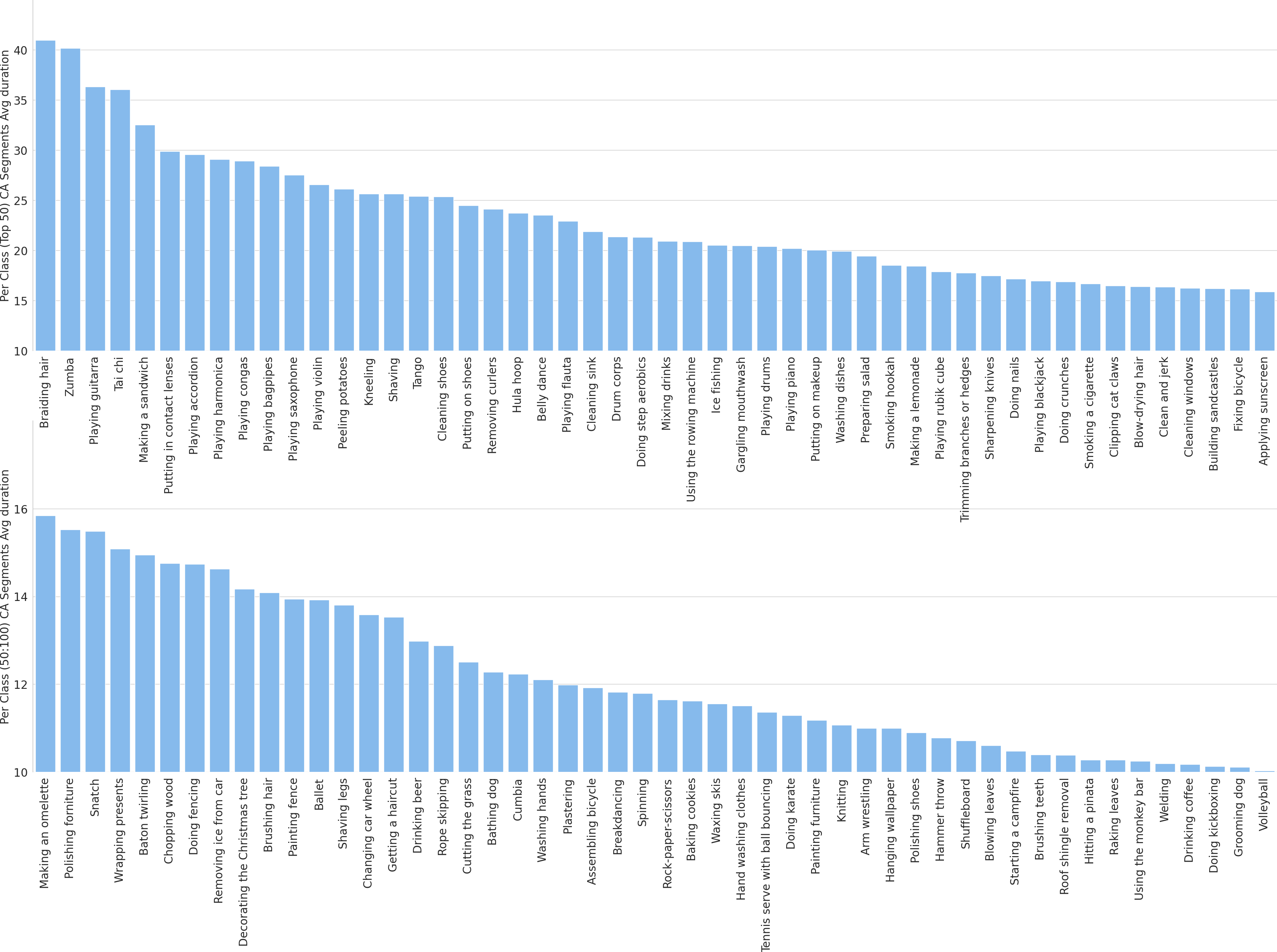}}
 \caption{(a) shows mean duration of each atomic action(top172), with three colors indicating person-person, person-object and pose atomic action types. (b) shows mean duration of complex activity. Due to limited space, we only show the 100 categories with the most instances.} 
 \label{fig:figure2}
\end{figure*}

\clearpage
{\small
\bibliographystyle{ieee_fullname}
\bibliography{egbib}
}